\documentclass[sigconf,nonacm]{acmart}

\usepackage{graphicx}
\usepackage{textcomp}
\usepackage{xcolor}
\usepackage{tabularx}
\usepackage{multirow}
\usepackage{threeparttable}
\usepackage{booktabs}
\usepackage{subcaption}
\usepackage{enumitem}
\usepackage{balance}
\usepackage[skip=1pt]{caption}

\newcommand{\modelname}{ChatTS}

\begin{document}

\title{ChatTS: Aligning Time Series with LLMs via Synthetic Data \\for Enhanced Understanding and Reasoning}

\author{Zhe Xie}
\affiliation{%
  \institution{Tsinghua University}
  \country{China}
  \city{Beijing}
}
\email{xiez22@mails.tsinghua.edu.cn}

\author{Zeyan Li}
\author{Xiao He}
\affiliation{%
  \institution{ByteDance}
  \country{China}
  \city{Beijing}
}

\author{Longlong Xu}
\affiliation{%
  \institution{Tsinghua University}
  \country{China}
  \city{Beijing}
}

\author{Xidao Wen}
\affiliation{%
  \institution{BizSeer}
  \country{China}
  \city{Beijing}
}

\author{Tieying Zhang}
\author{Jianjun Chen}
\affiliation{%
  \institution{ByteDance}
  \country{USA}
  \city{San Jose}
}

\author{Rui Shi}
\affiliation{%
  \institution{ByteDance}
  \country{China}
  \city{Beijing}
}

\author{Dan Pei}
\affiliation{%
  \institution{Tsinghua University}
  \country{China}
  \city{Beijing}
}

\begin{abstract}
Understanding time series is crucial for its application in real-world scenarios.
Recently, large language models (LLMs) have been increasingly applied to time series tasks, leveraging their strong language capabilities to enhance various applications.
However, research on multimodal LLMs (MLLMs) for time series understanding and reasoning remains limited, primarily due to the scarcity of high-quality datasets that align time series with textual information.
This paper introduces ChatTS, a novel MLLM designed for time series analysis. ChatTS treats time series as a modality, similar to how vision MLLMs process images, enabling it to perform both understanding and reasoning with time series.
To address the scarcity of training data, we propose an attribute-based method for generating synthetic time series with detailed attribute descriptions. 
We further introduce Time Series Evol-Instruct, a novel approach that generates diverse time series Q\&As, enhancing the model's reasoning capabilities.
To the best of our knowledge, ChatTS is the first MLLM that takes multivariate time series as input for understanding and reasoning, which is fine-tuned exclusively on synthetic datasets.
We evaluate its performance using benchmark datasets with real-world data, including six alignment tasks and four reasoning tasks. Our results show that ChatTS significantly outperforms existing vision-based MLLMs (e.g., GPT-4o) and text/agent-based LLMs, achieving a 46.0\% improvement in alignment tasks and a 25.8\% improvement in reasoning tasks.
We have open-sourced the source code, model checkpoint and datasets at \url{https://github.com/NetManAIOps/ChatTS}.
\end{abstract}

\maketitle

\section{Introduction}
\label{sec:introduction}
\begin{figure}[!t]
    \centering
    \includegraphics[width=\linewidth]{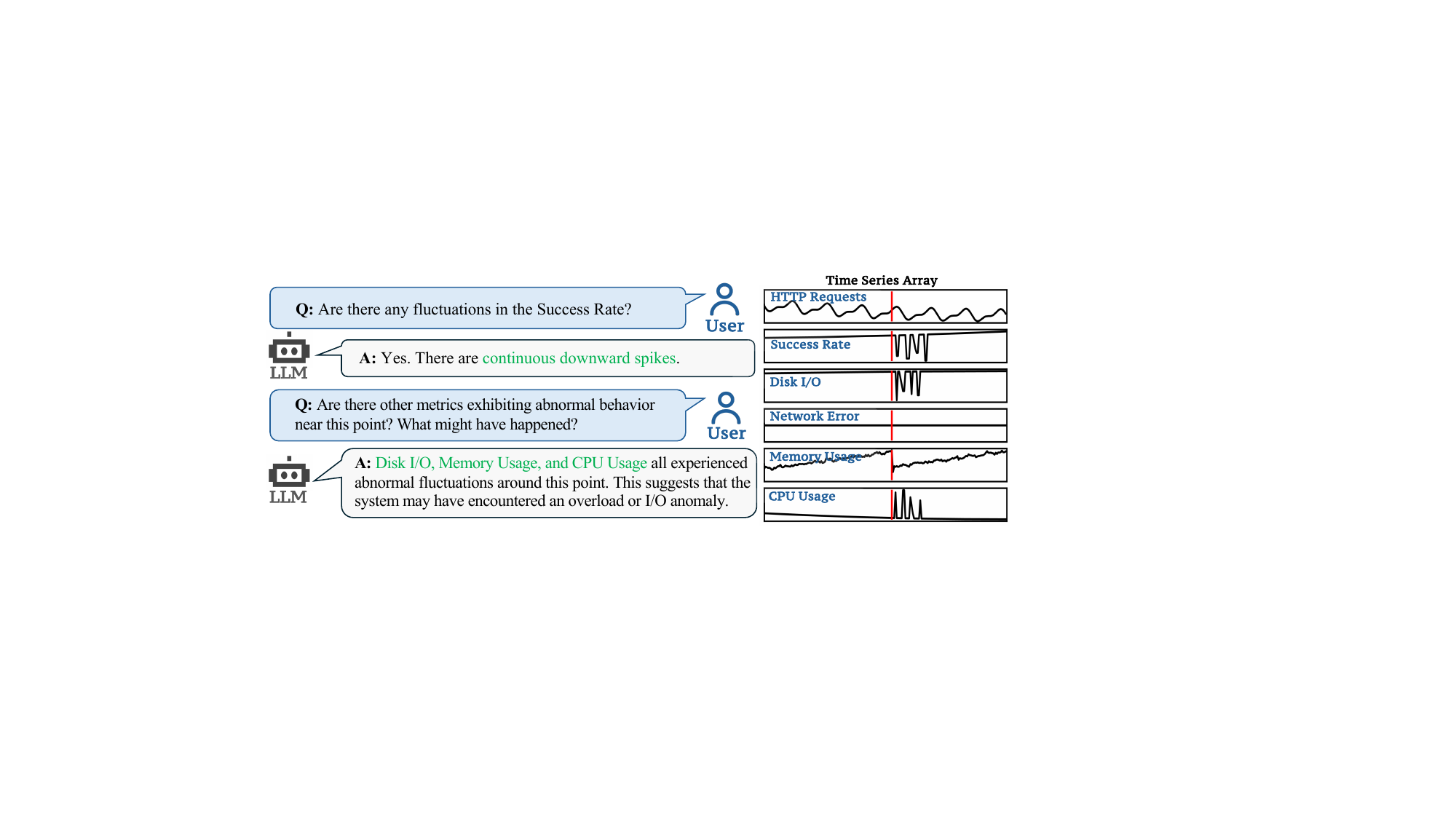}
    \setlength{\belowcaptionskip}{-10pt}
    \caption{Example of an AIOps application of time series-related dialogue.}
    \label{fig:tsqa_intro}
\end{figure}

Multimodal large language models (MLLMs) have recently achieved significant progress in vision-language tasks, showing exceptional performance even in scenarios requiring complex understanding and reasoning~\cite{liu2024visual,bai2023qwen,li2023blip,yin2024survey}. 
However, this success has not been replicated in the time series domain.
Even though some studies have attempted to integrate LLMs with time series, such as TimeLLM~\cite{jin2023time}, they usually only focus on specific classical time series tasks (e.g., forecasting) rather than understanding, reasoning, and dialogue based on time series attributes, as well as integrating into existing LLM workflows.
Moreover, recent studies indicate that LLMs still struggle with zero-shot reasoning about time series~\cite{merrill2024language}. 
This is particularly significant because time series analysis, widely applied in domains such as electricity \cite{wolde2006electricity}, healthcare \cite{penfold2013use}, traffic \cite{li2015trend}, weather \cite{lim2021time}, and finance \cite{sezer2020financial}, frequently requires understanding and reasoning about time series patterns.
Therefore, the ability to reason using both text and time series data is a critical capability for MLLMs, enabling them to support human decision-making by providing natural language explanations that align with human logic. 
Figure \ref{fig:tsqa_intro} illustrates such an example in an AIOps~\cite{zhong2023survey} scenario where understanding and reasoning about multivariate system monitoring time series are achieved through natural language dialogue, thereby improving the diagnostic and troubleshooting process.

Existing LLM-based methods for understanding and reasoning about time series attributes can be broadly categorized into text-based, vision-based, and agent-based approaches. Text-based methods directly use LLMs by structuring historical observations as raw text \cite{alnegheimish2024large}. However, these methods are often constrained by the limitation of prompt length and generally perform poorly in understanding the global features of time series compared to vision-based methods. Vision-based methods utilize vision MLLMs, which accept plot figures of time series data \cite{merrill2024language}, such as GPT-4o~\cite{gpt4o} or Qwen-VL~\cite{bai2023qwen}. While these methods can better capture global features, they are limited by the resolution of the plotted figures and face challenges in accurately interpreting the details. Recent works also show how agents can leverage time series analysis tools to interact with LLMs \cite{dbot, rcagent}. However, the ability of agents to understand time series is restricted by the functionality of the tools. 

\begin{figure*}[!t]
    \centering
    \includegraphics[width=0.95\linewidth]{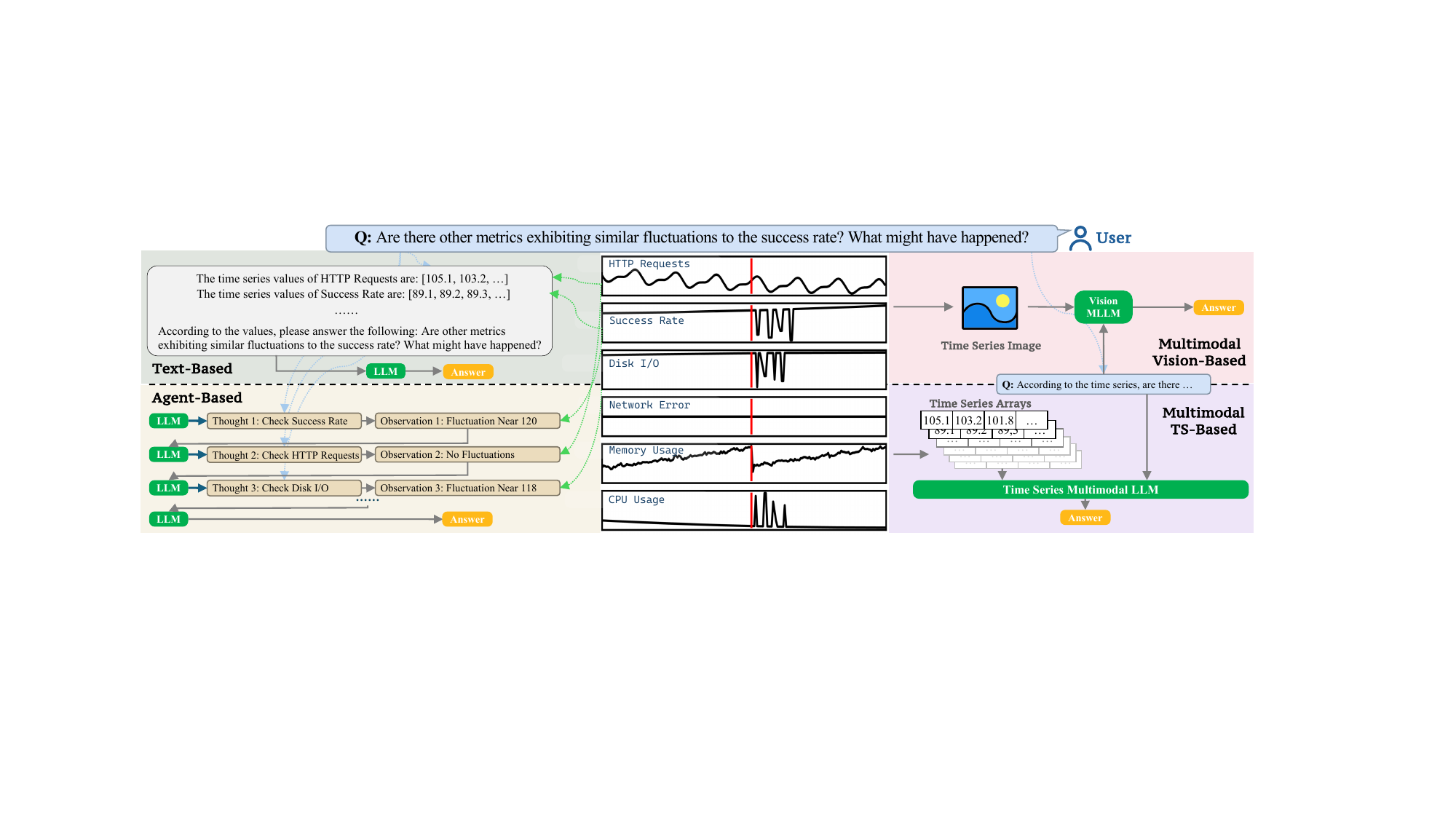}
    \caption{Comparison of four kinds of LLM-based methods for time series understanding and reasoning.}
    \label{fig:tsqa_methods}
\end{figure*}

Therefore, there is a strong need for TS-MLLM, a MLLM that can naturally handle time series modality, akin to how vision MLLMs process images.
Such models have the potential to unlock valuable insights from time series by providing intuitive, question-driven analysis capabilities.
Specifically, TS-MLLMs can capture global and local features and relationships between multivariate time series (MTS), areas where existing LLMs and MLLMs have struggled. By incorporating textual modalities as input, these models can broaden their applicability and better contextualize time series data, aligning the analysis with user queries. If successful, TS-MLLMs could perform novel tasks such as citing patterns and events in time series as evidence for observations and inferences, drawing interpretable conclusions from complex dynamical systems, and recognizing and responding to temporal patterns~\cite{merrill2024language}.

However, developing TS-MLLMs with effective understanding and reasoning ability for time series attributes faces several core challenges. 
First, multimodal time series data, especially language-time series pair data, is extremely scarce \cite{merrill2024language,chow2024towards,jin2024position}. Unlike modalities such as images and audio, almost no research focuses on the language alignment of time series. As a result, there is a significant lack of time-series + text data, which makes the construction of time-series dialogue and reasoning datasets challenging. This is fundamental for TS-MLLMs to develop temporal understanding and reasoning capabilities.
Second, time-series data contains abundant shape and numerical attributes (\textit{i.e.}, the types of local fluctuations and their amplitudes). Therefore, a diverse range of text is needed to comprehensively describe these attributes while ensuring accuracy to achieve effective alignment.
Third, real-world time-series data are usually variable in length, multivariate, and of uncertain quantity. The correlations among MTS are often a focus of attention (as illustrated in Figure \ref{fig:tsqa_intro}). In MLLMs for other modalities, such as images, few methods emphasize the relationships between multiple samples. However, such relationships are indispensable for understanding and reasoning about time series.
Finally, there is a lack of evaluation data and methods for TS-MLLMs. Developing comprehensive and reasonable datasets and methodologies to evaluate their performance is necessary.

To address the challenges above, we innovatively propose a method to fine-tune a pre-trained LLM for TS-MLLMs solely using synthetic time series and text data.
An important reason is that synthetic time series data for time series model training has shown good results~\cite{fu2024synthetic}. However, current methods are difficult to apply directly because time series-text alignment tasks require both \textit{precise} and \textit{diverse} time series attribute descriptions.
Therefore, we propose an attribute-based method for generating synthetic time series and precise text attributes to facilitate the modal alignment of time series with LLMs.
Compared with existing studies on synthetic time-series generation~\cite{fu2024synthetic,zhang2018generative}, the proposed attribute-based time-series generation method provides precise textual attributes for each detailed pattern of the time series, laying a foundation for generating diverse text data.
Furthermore, to equip MLLM with enhanced time series understanding and reasoning capabilities, we propose the Time Series Evol-Instruct (TSEvol) algorithm.
Through the diverse combinations of attributes and tasks, TSEvol can generate diverse time series Q\&A datasets through evolutions, thereby enhancing the model's overall performance.
To handle multivariate time-series inputs and fully preserve semantic information, we propose \modelname{}, trained using the generated synthetic datasets. \modelname{} employs a context-aware time-series encoder capable of encoding time series of (theoretically) arbitrary length and quantity while retaining their original numerical information.
Finally, to support comprehensive evaluation regarding both language alignment and time series reasoning, we have collected evaluation datasets comprising both real and synthetic time series. These datasets include both alignment and reasoning tasks with uni/multivariate time series, ensuring a thorough assessment of the model's performance.

\noindent\textbf{Our contributions.} This paper makes the following contributions.
\begin{itemize}[leftmargin=*]
    \item We propose to align LLMs with time series using attribute-based synthetic time series and text data. Building on this, we further introduce Time Series Evol-Instruct (TSEvol), an algorithm that generates diverse, accurate, and multimodal training datasets of time series and text entirely through synthetic data generation.
    \item We propose a context-aware TS-MLLM, \modelname{}, designed for variable-length, multivariate time series input and trained using the generated synthetic data. To the best of our knowledge, \modelname{} is \textit{the first TS-MLLM with multivariate time series as input for understanding and reasoning about time series attributes}.
    \item We have collected evaluation datasets containing real-world time series data, including six alignment tasks and four reasoning tasks. Evaluation results across multiple datasets demonstrate that \modelname{} significantly outperforms baseline models, including GPT-4o, in both time series alignment and reasoning tasks.
    \item We have open-sourced the model, source code, and evaluation datasets to support future research: \url{https://github.com/NetManAIOps/ChatTS}.
\end{itemize}

\section{Preliminary and Motivation}
\label{sec:motivation}
\subsection{Problem Definition}
The task of a TS-MLLM is to generate text-based responses based on the input textual query and MTS array.
Given a set of time series $\mathcal{T} = \{T_1, T_2, \dots, T_n\}$, where each $T_i = \{t_{i,1}, t_{i,2}, \dots, t_{i,m_i}\}$ represents a sequence of $m_i$ observed values over time for the $i$-th metric, and a natural language question $Q$, the goal is to generate an answer $A$ that captures relevant patterns or relationships across $\mathcal{T}$ based on the context of $Q$.
Formally, it can be defined as follows:

\begin{itemize}[leftmargin=*]
    \item \textbf{Input:} 
    \begin{itemize}
        \item A set of time series $\mathcal{T} = \{T_1, T_2, \dots, T_n\}$, where $T_i \in \mathbb{R}^{m_i}$ represents the values of the $i$-th metric over $m_i$ time points.
        \item A natural language query $Q$ specifies the information of interest within the time series data.
    \end{itemize}
    
    \item \textbf{Output:} 
    A text answer $A$ derived from the $\mathcal{T}$ analysis, providing insights based on $Q$.
\end{itemize}

The task of TS-MLLM can be expressed as a function:
\[
f(Q, \mathcal{T}) \rightarrow A,
\]
where $f$ denotes the model or algorithm responsible for interpreting the text query $Q$ and generating the text answer $A$ by analyzing relevant patterns and relationships across the time series in $\mathcal{T}$.

\subsection{Existing Methods}
Although mainstream LLMs currently do not support the direct input of time series modality data, time series information can be provided to LLMs through alternative methods to do simple understanding and reasoning about time series attributes, as shown in Figure \ref{fig:tsqa_methods}.
Existing approaches can be broadly categorized into text-based, vision-based, and agent-based, each with distinct limitations.

Text-based methods encode time series values as raw text~\cite{alnegheimish2024large}. However, these methods are constrained by the length of prompts, limiting their global analysis capabilities and often resulting in an incomplete understanding of the data context (refer to Section \ref{sec:evaluation}).

Vision-based approaches, which use visual representations of time series data (e.g., time series plots) processed by vision MLLMs~\cite{gpt4o,bai2023qwen}, may face challenges in accurately capturing detailed information in time series, resulting in lower accuracy for data-intensive tasks and high computational overhead (refer to Section \ref{sec:evaluation}).

Agent-based methods employ a reasoning and action strategy, breaking down complex tasks into a sequence of thoughts, observations, and actions conducted by external tools to analyze time series. While potentially more flexible, this approach is heavily dependent on expert knowledge and effectiveness of tools, token-intensive, and time-consuming, often requiring extensive token chains to handle MTS data. Additionally, hallucination becomes a significant problem~\cite{yoffe2024debunc} as the chains grow longer, reducing reliability in complex analytical tasks.

\subsection{Time Series Multimodal LLM}
TS-MLLM is a new type of MLLM that aims at overcoming the limitations of existing methods by \textit{natively} integrating both textual and time series inputs (see Figure \ref{fig:tsqa_methods}).
It can process multiple time series data and textual descriptions, enabling a unified analysis that captures complex, multivariate relationships.
Unlike previous methods, it does not rely on lengthy token chains or visual representations, thereby reducing computational overhead and mitigating issues with hallucination.
Through the alignment of time series and text, TS-MLLM can perform both global and local analysis of the shape and numerical information of time series. This capability allows it to achieve higher accuracy and greater potential than existing methods.

\section{Methodology}
\label{sec:method}
\subsection{Overview}
\begin{figure}[!htbp]
    \centering
    \includegraphics[width=\linewidth]{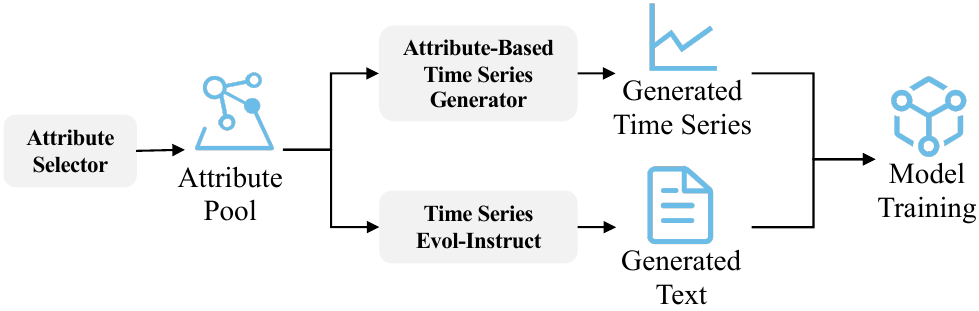}
    \setlength{\belowcaptionskip}{-10pt}
    \caption{Overview of \modelname{}.}
    \label{fig:overview}
\end{figure}
Due to the scarcity of high-quality datasets that align time series with textual information, we propose to generate synthetic text-time series pairs for model training.
Synthetic data is a common approach when there is a lack of sufficient real training data, and its effectiveness has been well validated in various fields~\cite{fu2024synthetic,savage2023synthetic,luo2023time}. However, as discussed earlier, "time series + text" data for TS-MLLM requires sufficient accuracy to ensure alignment precision, comprehensive coverage of time series attributes to guarantee effective multimodal alignment, and task diversity in the text to enhance QA and reasoning abilities. Unfortunately, existing time series generation methods~\cite{fu2024synthetic,zhang2018generative} fail to achieve these goals. A key reason is that we need a \textit{diverse} set of time series and \textit{precise, detailed} descriptions of time series patterns. Therefore, in this paper, we propose an attribute-based method to generate time series + text data, as illustrated in Figure \ref{fig:overview}:
\begin{figure*}[!htb]
    \centering
    \includegraphics[width=1.0\linewidth]{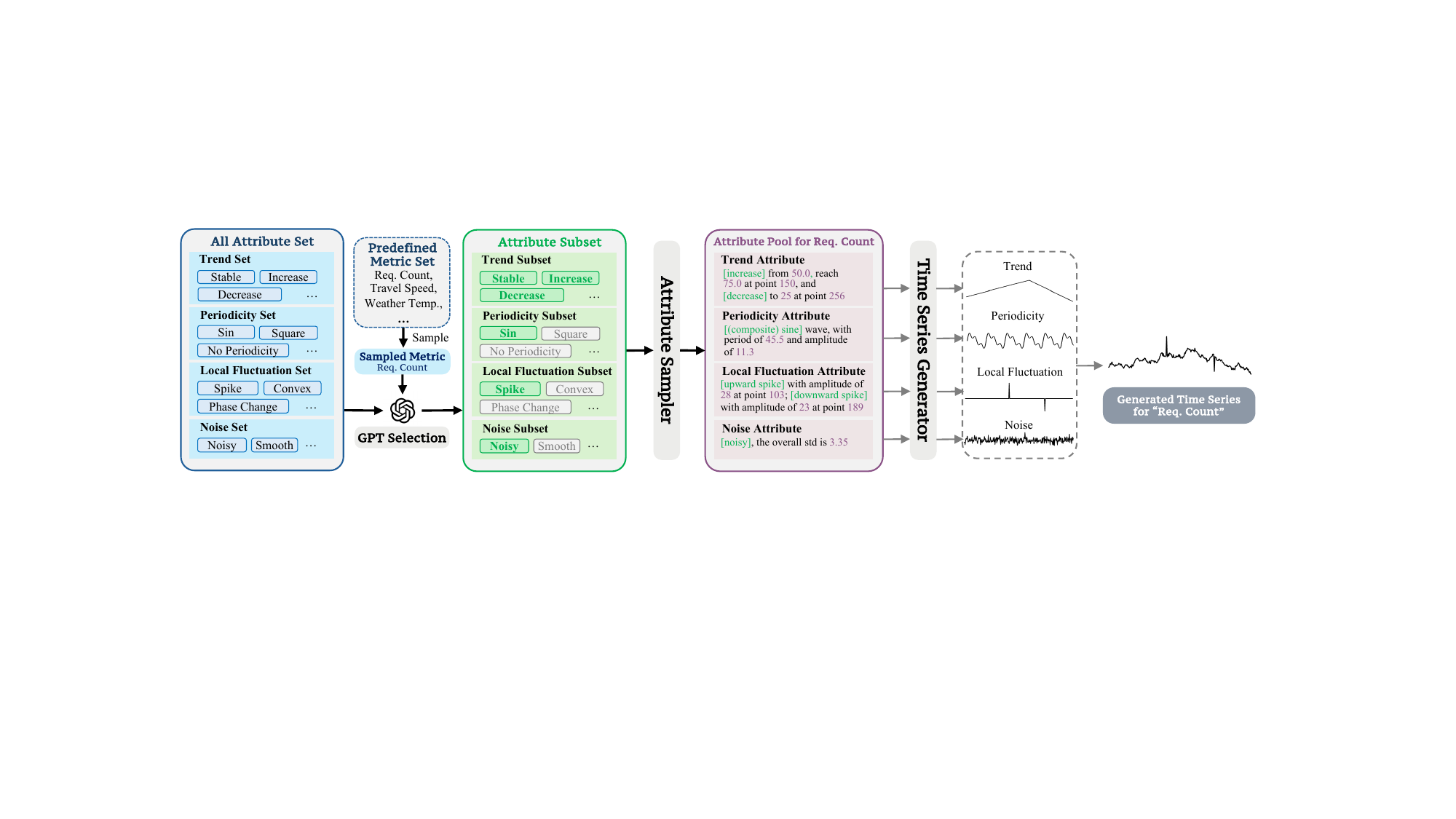}
    \setlength{\belowcaptionskip}{-10pt}
    \caption{Attribute selector and attribute-based time series generator in \modelname{}.}
    \label{fig:feature_generator}
\end{figure*}

\begin{itemize}[leftmargin=*]
    \item \textbf{Attribute Selector} (Section \ref{sec:time_series_generator}): To produce highly controllable time-series data with precise attributes, we use a detailed feature set to describe time series. These attributes are aligned with real-world settings through an LLM selection.
    \item \textbf{Attribute-Based Time Series Generator} (Section \ref{sec:time_series_generator}):
    Construct time series that correspond exactly to the attribute pool using a rule-based approach.
    \item \textbf{Time Series Evol-Instruct} (Section \ref{sec:evol_instruct}): A novel Time Series Evol-Instruct module for creating large, diverse, and accurate datasets of time-series and text question-answering pairs for complex reasoning.
    \item \textbf{Model Design} (Section \ref{sec:mllm}): To handle MTS, we design a context-aware MLLM encoding for multiple time series input, along with a value-preserved time series encoding method.  
    \item \textbf{Model Training} (Section \ref{sec:model_training}): A large-scale training and a SFT are conducted to perform language alignment and improve time series-related reasoning ability.
\end{itemize} 
As shown in Figure \ref{fig:overview}, the framework in \modelname{} integrates synthetic data generation and model training into a pipeline that ensures effective time series attributes understanding and reasoning with only synthetic data. First, building on the attribute-based time series generation and TSEvol described in Sections 3.2 and 3.3, the pipeline generates synthetic data that captures intricate numerical and textual information for effective multimodal alignment. This data is then used to train the model (Section 3.4), where the context-aware time series encoder preserves the time series values while aligning attributes with textual semantics accurately.
Finally, by alignment training and SFT (Section 3.5), \modelname{} achieves precise alignment between time series encoding and text embeddings, along with enhanced reasoning capabilities.

\subsection{Attribute-Based Time Series Generator}
\label{sec:time_series_generator}

Diverse time series and precise, detailed textual attribute descriptions are essential to achieve accurate time series language alignment.
Time series have rich pattern attributes, which can be roughly categorized into trend, periodicity, and remainder~\cite{rb1990stl, DBLP:journals/pvldb/HeLTWL23}. Much existing research on the generation of time series~\cite{fu2024synthetic,fons2024evaluating} also adopts similar approaches to classify these attributes. Therefore, following existing studies, we classify time series attributes into four major categories, Trend, Periodicity, Noise, and Local Fluctuation, to construct the corresponding attribute set for time series.

Based on this, we propose an attribute selector and an attribute-based time series generator that produces synthetic time series data (see Figure \ref{fig:feature_generator}).
First, we define an ``All Attribute Set'', which includes many specific attributes under different attribute categories.
The All Attribute Set includes 4 types of Trend, 7 types of Seasonality, 3 types of Noise, and 19 types of local fluctuations. The complete list can be found in the source code.
Different attributes within the same category can be combined. A time series can include multiple segments of trends and several local fluctuations by combining the same type of attributes (see Figure \ref{fig:feature_generator}). Additionally, by combining sine waves, we can generate a diverse range of periodic fluctuation patterns. Therefore, the proposed time series generator can theoretically generate an infinite number of different time series, ensuring the richness of attributes.
We also introduced a GPT Selector. Specifically, when generating an attribute set for time series, we randomly sample a metric from a large ``Metric Set'' that contains 567 predefined metric names from real-world scenarios and use GPT to choose a \textit{attribute subset} from the all attribute set, based on the actual physical meaning of the metric and the predefined scenario. This helps align time series with real-world physical meanings.

Then, the \textit{Attribute Sampler} randomly samples a combination of attributes from the Attribute Subset. It also assigns specific numerical values, like position and amplitude, based on rules and constraints from the GPT Selector. These details are stored in the ``Attribute Pool'', which records all the detailed information about a time series. The \textit{Time Series Generator} finally creates time series arrays that \textit{exactly} match the attributes from the pool in a rule-based manner (more details can be found in the source code). This process allows us to generate diverse synthetic time series with precise attribute descriptions.

\subsection{Time Series Evol-Instruct}
\label{sec:evol_instruct}
\begin{figure}[!htbp]
    \centering
    \includegraphics[width=\linewidth]{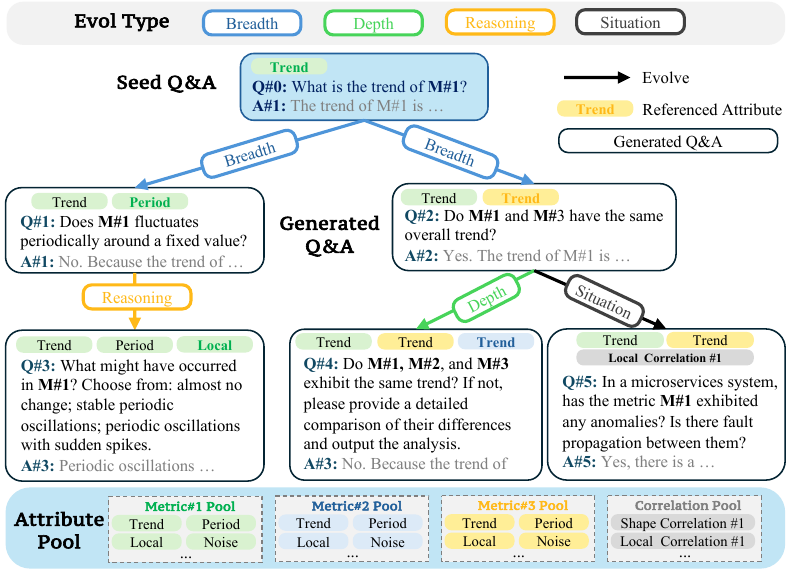}
    \setlength{\belowcaptionskip}{-10pt}
    \caption{Time Series Evol-Instruct}
    \label{fig:evol_instruct}
\end{figure}

\begin{figure*}[!t]
    \centering
    \includegraphics[width=0.9\linewidth]{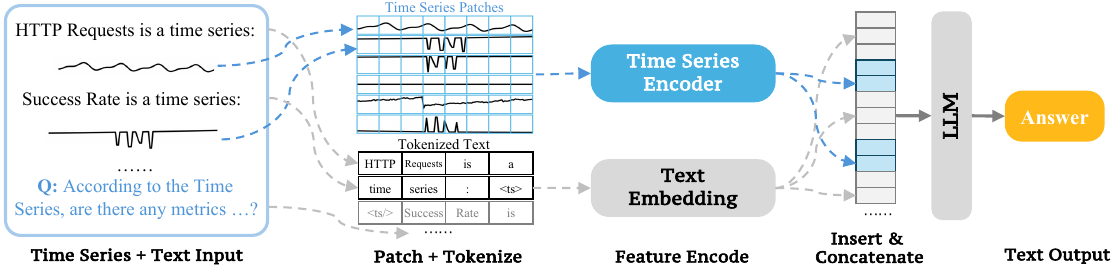}
    \setlength{\belowcaptionskip}{-5pt}
    \caption{Model Structure of the Multimodal LLM in \modelname{}}
    \label{fig:model}
\end{figure*}

To improve the model's question-answering and reasoning abilities, it is essential to have high-quality SFT training data that is diverse in format and tasks. However, due to the lack of time-series + text data, it is challenging to obtain sufficiently diverse time-series-related training data directly. To generate accurate time-series + text SFT data with rich question-answering formats, inspired by Evol-Instruct~\cite{xu2023wizardlm} and its multimodal version MMEvol~\cite{luo2024mmevol}, we innovatively propose Time Series Evol-Instruct (TSEvol).

Evol-Instruct~\cite{xu2023wizardlm} is a data generation approach that incrementally evolves instructional prompts and their outputs to enhance the diversity and complexity of training datasets for LLMs.
TSEvol builds upon Evol-Instruct by introducing a mechanism to incorporate time series attributes dynamically into each evolutionary step (see Figure \ref{fig:evol_instruct}).
TSEvol relies on \textit{attribute pools} of multivariate time series (see Section \ref{sec:time_series_generator}). Additionally, to enhance the model's ability to analyze correlations, we introduce a correlation pool, which records time series with related attributes (refer to the source code for details).
During each step of the evolution process, a subset of attributes is randomly selected from the \textit{attribute pool} and added as \textit{additional context}, guiding the LLMs to generate Q\&As about a broader set of time series attributes according to the \textit{evolution type}.
With TSEvol, generated Q\&As can cover more attributes in the time series and avoid repetitive questions. We also added an attribute-based eliminator to ensure the Q\&As match the time series attributes.
In addition to the commonly used evolution types, we also add two more types, reasoning (reasoning-based questions) and situation (situation-based questions), to enhance the model's ability to handle complex questions.

\subsection{Time Series Multimodal LLM}
\label{sec:mllm}
In this subsection, we introduce the model structure of the proposed \modelname{}, as shown in Figure \ref{fig:model}. \modelname{} takes multivariate time series and text, along with their \textit{contextual information} as the input.
\subsubsection{Context-Aware Time-Series Multimodal LLM}
To handle the multimodal inputs, \modelname{} first separates the input time series arrays and the text.
Following the established practice in encoding time series for LLMs~\cite{jin2023time}, the input time series arrays are divided into fixed-size patches, which enables the model to handle and encode temporal patterns more effectively.
We employ a simple 5-layer MLP to encode each patch of the time series, as time series inherently have sequential patterns. Therefore, a simple structure can map the patch features to a space aligned with the text embedding.
For text input, they are tokenized and then encoded through a text embedding layer.
In this way, each patch of the time series and each text token are mapped to the same space.

To fully retain the contextual information of multivariate time series, we performed token-level concatenation based on the position of the time series in the original input. Specifically, the encoded patches corresponding to each time series were inserted between the surrounding text tokens. Unlike the method used in TimeLLM~\cite{jin2023time}, this approach ensures that the contextual information of the time series is fully preserved. This is especially important in multivariate scenarios, where referencing the corresponding time series in textual form is often necessary.
This process results in a sequence that reflects the multivariate structure of the data, enabling the LLM to capture both temporal and contextual dependencies across different metrics. This sequence is then fed into the LLM, which generates an answer that incorporates insights from both the time series data and the natural language query, achieving a multimodal understanding suited for complex question-answering tasks.

\subsubsection{Value-Preserved Time Series Normalization}
\begin{figure}[!htbp]
    \centering
    \includegraphics[width=\linewidth]{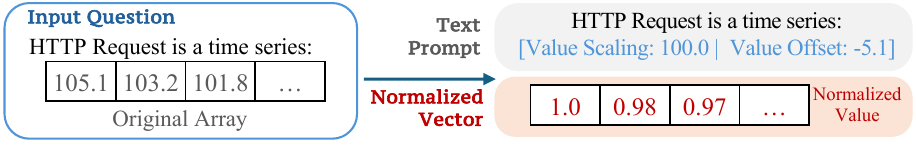}
    \setlength{\belowcaptionskip}{-10pt}
    \caption{Value-Preserved Time Series Normalization}
    \label{fig:encode}
\end{figure}

The numerical features of time series are essential, as real-world applications often involve specific numerical queries (e.g., asking for the maximum CPU utilization). However, normalization of time series data can lead to losing original numerical information. To address this, we introduce a value-preserved time series normalization scheme (as shown in Figure \ref{fig:encode}).
First, we apply standard min-max normalization (0-1 scaling) to each time series array. 
Then, for each time series, we include the normalization parameters-``Value Scaling'' (the scaling factor during normalization) and ``Value Offset'' (the offset applied during normalization)—in the text \textbf{as part of a prompt}. 
This approach leverages the numerical understanding capabilities of LLMs, enabling us to normalize time series features while preserving the original numerical information.
To further enhance numerical understanding, numerical tasks are included in the training dataset (see Section \ref{sec:model_training}).

\begin{figure*}[!t]
    \centering
    \includegraphics[width=0.8\linewidth]{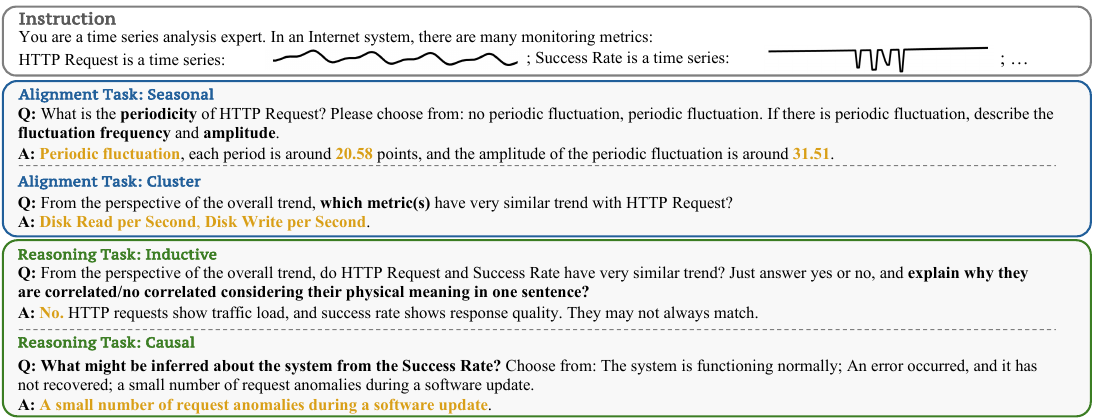}
    \setlength{\belowcaptionskip}{-5pt}
    \caption{Example QAs in some evaluation tasks.}
    \label{fig:evaluation_examples}
\end{figure*}

\subsection{Model Training}
\label{sec:model_training}
{
\footnotesize
\begin{table}[!htbp]
    \centering
    \caption{Training Datasets}
    \label{tab:training_datasets}
    \begin{tabular}{c|ccc|cc}
        \toprule
        \textbf{Stage} & \multicolumn{3}{c}{\textbf{Alignment}} & \multicolumn{2}{c}{\textbf{SFT}} \\
        \cmidrule(lr){2-4}\cmidrule(lr){5-6}
        Dataset & UTS & MTS-Shape & MTS-Local & TSEvol & Instruct Follow \\
        \midrule
        \textbf{\# Samples} & 35,000 & 35,000 & 35,000 & 24,270 & 5,050 \\
        \bottomrule
    \end{tabular}
    
\end{table}
}

\modelname{} is trained based on QWen2.5-14B-Instruct~\cite{yang2024qwen2}\footnote{https://huggingface.co/Qwen/Qwen2.5-14B-Instruct}, with a two-stage fine-tuning process: large-scale alignment training and supervised fine-tuning (SFT).
Table \ref{tab:training_datasets} shows the datasets we use during training.

\subsubsection{Large-Scale Alignment Training}
In the first stage, we perform large-scale alignment training using the attribute-based synthetic time series data to establish an initial alignment between the text and time series modalities within the LLM. This stage enables \modelname{} to align textual descriptions with time series attributes effectively.
During the alignment stage, we created three datasets for large-scale training based on a series of manually designed templates and LLM refinement. The \textit{UTS} dataset includes tasks for basic attribute descriptions of univariate time series (both global and local attribute tasks are included). The \textit{MTS-Shape} dataset consists of multivariate data with \textit{global} trend correlations designed to enhance the model's ability to analyze multivariate correlations. The \textit{MTS-Local} dataset contains multivariate data with correlated \textit{local} fluctuations, aiming to improve the model's capability in analyzing local features of multivariate data.
Given MTS’s more complex feature combinations, we set the training data size for MTS and UTS at an approximately 2:1 ratio. We conduct a dataset scaling study in Section \ref{sec:dataset_scaling} to investigate the impact of training dataset size.

\subsubsection{Supervised Fine-Tuning}
In the second stage, we use SFT to develop the LLM’s ability to perform complex question-answering and reasoning tasks. 
This stage utilizes two main types of training data: the datasets generated with TSEvol, designed to enhance the model’s question answering and reasoning ability about time series, and an instruction-following (IF) dataset, constructed based on a series of predefined templates, designed to enhance the model's ability to follow specific response formats.
For TSEvol, we used the dataset from alignment training along with LLM-generated QAs as the seed data.
Together, these datasets train the multimodal LLM to respond accurately to time series-specific queries and follow task instructions, strengthening its capacity for complex, context-driven question-answering and reasoning tasks. 
In both Alignment and SFT stages, we enhance ChatTS’s numerical capabilities through a series of numerical tasks. Specifically, we explicitly train the model to learn various aspects, such as maximum/minimum values, segmented averages, local features (e.g., spike positions and amplitudes), seasonality and trend amplitudes, and raw numerical values at individual time points. The numerical evaluation metrics in our experimental results further demonstrate ChatTS’s strong performance in time series numerical analysis.

\subsubsection{Training Settings}
We use QA pairs as the data format for both training stages.
During alignment training, we mixed in a small amount of IF data and found that this mitigates the decline in the model's IF ability. In the SFT stage, we mixed 30\% of the alignment training dataset to reduce overfitting. The training dataset includes time series with lengths ranging from 64 to 1024 to ensure that \modelname{} can handle varying time series lengths.
Full-parameter SFT is used for \modelname{} with DeepSpeed~\cite{deepspeed} and LLaMA-Factory~\cite{zheng2024llamafactory}, with Qwen2.5-14B-Instruct~\cite{yang2024qwen2,qwen14b} as the base model. 
Inference for both Qwen and \modelname{} is also conducted with DeepSpeed.

\section{Evaluation}
\label{sec:evaluation}
In this section, we will comprehensively evaluate the performance of \modelname{} by answering the following research questions (RQs):
\begin{itemize}[leftmargin=*]
    \item \textbf{RQ1.} How well does \modelname{} align with time series?
    \item \textbf{RQ2.} How does \modelname{} perform in time series reasoning tasks?
    \item \textbf{RQ3.} Are attribute-based data and TSEvol effective?
    \item \textbf{RQ4.} How does the training set size affect model performance?
    \item \textbf{RQ5.} Is the time series modalilty in \modelname{} truly useful?
    \item \textbf{RQ6.} Does \modelname{}, with its native time-series multimodal capabilities, have advantages over agent-based methods?
\end{itemize}

\subsection{Experimental Setup}
\subsubsection{Evaluation Tasks}
To comprehensively evaluate the model's performance, we set two categories of evaluation tasks: alignment tasks and reasoning tasks, following the general evaluation methods of multimodal LLMs~\cite{liu2024visual,luo2024mmevol,cai2024timeseriesexam}.
For each type of evaluation task, we designed a series of subtasks based on existing work. Some example QAs are shown in Figure \ref{fig:evaluation_examples} (more details can be found in the source code). Specific tasks that rely heavily on domain-specific knowledge (e.g., classification and etiological reasoning) were excluded due to the lack of high-quality datasets that provide sufficient background information. Therefore, we primarily focused on the following tasks:

Alignment tasks are divided into univariate and multivariate:
\begin{itemize}[leftmargin=*]
    \item \textbf{Univariate tasks.} Identify trends, seasonality, noise, and local fluctuations. These tasks include both \textit{categorical} subtasks and \textit{numerical} subtasks.
    \item \textbf{Multivariate tasks.} Correlation and clustering. These tasks are all categorical.
\end{itemize}

The reasoning tasks include inductive reasoning, deductive reasoning, causal reasoning, and comparison reasoning (MCQ2):
\begin{itemize}[leftmargin=*]
    \item \textbf{Inductive reasoning.} Q\&A task. Inductive summarization of the physical meaning reflected by a uni/multivariate time series.
    \item \textbf{Deductive reasoning.} True/False (T/F) task. Reasoning based on a predefined condition in conjunction with univariate time series.
    \item \textbf{Causal reasoning.} Multiple-choice task. Based on univariate time series, select the most likely cause.
    \item \textbf{Comparison reasoning (MCQ2).} Multiple-choice task. Compare two time series and select the correct answer.
\end{itemize}
More details about the evaluation tasks can be found in the source code and the evaluation dataset.

\subsubsection{Evaluation Metrics}
For categorical tasks in alignment evaluation, we match labels from the responses of LLMs using rule-based matching and use F1-Score as the metric.
For numerical tasks in alignment evaluation, we extract numbers from the responses of LLMs and use \textit{relative accuracy} (1.0 - relative error) as the metric:
\begin{equation*}
    relative\_accuracy = \max\left(1.0 - \frac{\left| V_{answer} - V_{label} \right|}{\left| V_{label} \right|}, 0.0 \right)
\end{equation*}
We set a minimum value of 0.0 for relative accuracy to mitigate the impact of outlier results.
For Q\&A tasks in inductive reasoning, answers are evaluated using RAGAS~\cite{es2023ragas}, a keyword-matching approach through LLM-based fuzzy matching.
T/F and MC tasks are directly evaluated through choice matching and the accuracy is calculated.
All evaluation metrics are the higher, the better.

\subsubsection{Evaluation Datasets}
{
\small
\begin{table}[!htbp]
    \centering
    \setlength{\abovecaptionskip}{2pt}
    \caption{Tasks in Evaluation Dataset}
    \label{tab:tasks}
    \begin{tabularx}{\linewidth}{>{\centering\arraybackslash}p{0.9cm}|>{\centering\arraybackslash}X>{\centering\arraybackslash}p{1.6cm}}
        \toprule
        \textbf{Dataset} & Tasks & \# Questions \\
        \midrule
        \multirow{3}{*}{\textbf{A}} & Alignment (Trend, Season, Noise, Local, Correlation, Cluster), Reasoning (Inductive, Deductive, Causal) & \multirow{3}{*}{525} \\
        \midrule
        \multirow{2}{*}{\textbf{B}} & Alignment (Trend, Season, Noise, Local, Correlation, Cluster), Reasoning (Inductive) & \multirow{2}{*}{1,616} \\
        \midrule
        \textbf{MCQ2} & Reasoning (Comparison - MCQ2) & 100 \\
        \bottomrule
    \end{tabularx}
\end{table}
}

Our evaluation is conducted on three datasets (see Table \ref{tab:tasks}) to test the model’s performance across both real-world and synthetic time series scenarios. Dataset A and B are collected by us, and Dataset MCQ2 is an open-source dataset~\cite{merrill2024language}.

\textit{Dataset A} includes real-world time series data collected from multiple domains, including AIOps~\cite{li2022constructing}, weather~\cite{weatherdataset}, the NAB (Numenta Anomaly Benchmark)~\cite{ahmad2017unsupervised}, and Oracle system metrics~\cite{li2022actionable}.
We manually label and collect a total of 525 questions, including both alignment tasks and reasoning tasks.

To expand the size of the evaluation set, we used the attribute-based time series generator introduced in \modelname{} to generate a series of time series and created alignment Q\&A by applying a set of templates. We also develop a set of reasoning questions with LLM, resulting in a larger-scale \textit{Dataset B} containing 1,616 questions. Considering the complexity of reasoning tasks, we have included only inductive reasoning tasks in the reasoning tasks of this dataset to ensure the quality of the questions.

\textit{MCQ2}~\cite{merrill2024language} is an open-source dataset~\cite{mcq2dataset} that includes comparison reasoning tasks.
The questions, answers, and time series in this dataset are all generated by LLMs. We did not use the etiological reasoning and forecasting datasets as they are not aligned with our evaluation settings. Furthermore, \cite{merrill2024language} suggests that the settings of the MCQ1 dataset are unsuitable for evaluating the performance of time series reasoning, so we also did not adopt it. Considering the inference cost, we randomly sampled 100 questions.

{
\footnotesize
\begin{table*}[t]
    \centering
    \caption{Comparison of different models in terms of performance and cost of input tokens on alignment tasks (*image tokens are converted in some models according to price). ``Cate.'' and ``Num.'' denotes categorical and numerical tasks respectively. F1-Score and relative accuracy are used in evaluating categorical and numerical tasks, respectively.}
    \label{tab:align}
    \begin{tabular}{c|cl|cccccccc|cc|cc|cc}
        \toprule
        \textbf{Dataset} & \textbf{Type} & \textbf{Model} & \multicolumn{2}{c}{Trend} & \multicolumn{2}{c}{Season} & \multicolumn{2}{c}{Noise} & \multicolumn{2}{c}{Local} & Corr. & Clus. & \multicolumn{2}{c}{\textbf{Overall}} & \textbf{Tokens} & \textbf{Est. Cost} \\
        \cmidrule(lr){4-5}\cmidrule(lr){6-7}\cmidrule(lr){8-9}\cmidrule(lr){10-11}\cmidrule(lr){12-12}\cmidrule(lr){13-13}\cmidrule(lr){14-15}
        & & Task & Cate. & Num. & Cate. & Num. & Cate. & Num. & Cate. & Num. & Cate. & Cate. & Cate. & Num. &  & \$ \\
        \midrule
        \multirow{10}{*}{\textbf{A}} & \multirow{4}{*}{Text} & GPT-4o-mini & 0.585 & 0.752 & 0.649 & 0.264 & \textbf{0.952} & 0.312 & 0.263 & 0.187 & 0.357 & 0.254 & 0.464 & 0.310 & 1.3M & 0.20 \\
        & & GPT-4o & 0.585 & \textbf{0.882} & 0.811 & 0.768 & 0.905 & 0.153 & 0.379 & 0.256 & 0.476 & 0.333 & 0.542 & 0.371 & 1.3M & 3.25 \\
        & & GPT-4-Turbo & 0.526 & 0.699 & 0.649 & 0.131 & 0.900 & 0.339 & 0.303 & 0.247 & 0.417 & 0.269 & 0.490 & 0.353 & 1.3M & 13.0 \\
        & & QWen2.5-14B & 0.707 & 0.709 & 0.622 & 0.205 & 0.833 & 0.231 & 0.137 & 0.099 & 0.571 & 0.349 & 0.464 & 0.241 & 1.3M & 0.35 \\
        \cmidrule(lr){2-17}
        & \multirow{2}{*}{Vision} & GPT-4o-mini & 0.610 & 0.501 & 0.432 & 0.205 & 0.667 & 0.201 & 0.242 & 0.184 & 0.357 & 0.330 & 0.404 & 0.248 & 2.2M* & 0.33 \\
        & & GPT-4o & 0.659 & 0.613 & 0.811 & 0.559 & 0.810 & 0.248 & 0.537 & 0.414 & 0.476 & 0.480 & 0.609 & 0.436 & 0.13M* & 0.32 \\
        \cmidrule(lr){2-17}
        & \multirow{2}{*}{Agent} & GPT-4o-mini & 0.559 & 0.773 & 0.595 & 0.270 & 0.714 & 0.105 & 0.400 & {0.212} & {0.381} & {0.361} & {0.469} & {0.309} & {3.0M} & {0.45} \\
        &  & {GPT-4o} & {0.537} & {0.650} & {0.405} & {0.000} & {0.595} & {0.088} & {0.232} & {0.136} & {0.429} & {0.417} & {0.390} & {0.220} & {2.7M} & {6.75} \\
        \cmidrule(lr){2-17}
        & \multirow{1}{*}{TS} & \textbf{\modelname{}} & \textbf{0.927} & 0.874 & \textbf{0.973} & \textbf{0.849} & 0.857 & \textbf{0.511} & \textbf{0.895} & \textbf{0.805} & \textbf{0.905} & \textbf{0.782} & \textbf{0.889} & \textbf{0.788} & \textbf{0.08M} & \textbf{0.02} \\
        \midrule
        \multirow{10}{*}{\textbf{B}} & \multirow{4}{*}{Text} & GPT-4o-mini & 0.619 & 0.716 & 0.711 & 0.317 & 0.427 & 0.198 & 0.145 & 0.091 & 0.335 & 0.269 & 0.336 & 0.217 & 4.5M & 0.67 \\
        & & GPT-4o & 0.690 & 0.825 & 0.732 & 0.474 & 0.573 & 0.331 & 0.191 & 0.136 & 0.324 & 0.281 & 0.366 & 0.284 & 4.5M & 11.3 \\
        & & GPT-4-Turbo & 0.667 & 0.732 & 0.667 & 0.345 & 0.348 & 0.067 & 0.188 & 0.133 & 0.438 & 0.369 & 0.385 & 0.259 & 4.5M & 45.0 \\
        & & QWen2.5-14B & 0.711 & 0.669 & 0.705 & 0.217 & 0.256 & 0.094 & 0.111 & 0.082 & 0.402 & 0.276 & 0.339 & 0.193 & 4.5M & 1.22 \\
        \cmidrule(lr){2-17}
        & \multirow{2}{*}{Vision} & GPT-4o-mini & 0.679 & 0.240 & 0.814 & 0.453 & 0.305 & 0.238 & 0.141 & 0.081 & 0.327 & 0.307 & 0.347 & 0.142 & 11.4M* & 1.71 \\
        & & GPT-4o & 0.702 & 0.361 & 0.938 & 0.589 & 0.610 & 0.398 & 0.375 & 0.265 & 0.367 & 0.389 & 0.472 & 0.311 & 0.56M* & 1.40 \\
        \cmidrule(lr){2-17}
        & \multirow{2}{*}{{Agent}} & {GPT-4o-mini} & {0.612} & {0.591} & {0.455} & {0.605} & {0.375} & {0.000} & {0.043} & {0.022} & {0.654} & {0.585} & {0.372} & {0.125} & {8.5M} & {1.27} \\
        &  & {GPT-4o} & {0.532} & {0.586} & {0.619} & {0.658} & {0.391} & {0.262} & {0.551} & {0.287} & {0.500} & {0.464} & {0.490} & {0.370} & {7.2M} & {10.8} \\
        \cmidrule(lr){2-17}
        & \multirow{1}{*}{TS} & \textbf{\modelname{}} & \textbf{0.976} & \textbf{0.902} & \textbf{1.000} & \textbf{0.930} & \textbf{0.927} & \textbf{0.572} & \textbf{0.828} & \textbf{0.752} & \textbf{0.818} & \textbf{0.834} & \textbf{0.862} & \textbf{0.787} & \textbf{0.34M} & \textbf{0.09} \\
        \bottomrule
    \end{tabular}
\end{table*}
}

\subsubsection{Baselines}
Based on different modalities, we categorized the baseline methods into the following types:

\begin{itemize}[leftmargin=*]
    \item \textbf{Text-Based:} These methods convert time series arrays into textual prompts as inputs for LLMs. We choose several mainstream LLMs as our base model (GPT-4o/GPT-4o-mini/GPT-4-Turbo/QWen2.5-14B-Instruct) for evaluation.

    \item \textbf{Vision-Based:} These methods plot time series and input them into visual MLLMs. We choose mainstream vision MLLMs (GPT-4o/GPT-4o-mini) for evaluation.

    \item \textbf{Agent-Based:} These methods employ the ReAct~\cite{yao2022react} framework to interact with multiple tools to analyze the time series. The tools used include single-point/range query, STL decomposition, anomaly detection (autoregression AD in adtk~\cite{adtk}), and classification (Rocket~\cite{dempster2020rocket}) for UTS; trend/fluctuation correlation (based on Pearson correlation \& rules), multivariate version of AD and classification for MTS. We choose GPT-4o/GPT-4o-mini for the agent. More details about the tools' implementation can be found in the source code.
    \textit{We also conducted additional experiments to explore further the capabilities of agent-based methods (Section \ref{sec:agent_study}), which studies the impact of tool accuracy}.
\end{itemize}

\subsubsection{Implementation}
For GPT-based models, we used OpenAI’s API to infer and track token consumption. For \modelname{} and QWen-based models, the training and inference are conducted locally on 8$\times$A800 GPUs. The token consumption for \modelname{} is calculated after the ``Reorder \& Concat'' step.

\subsection{RQ1. Alignment Tasks}
The evaluation results on alignment tasks are shown in Table~\ref{tab:align}. ChatTS consistently outperforms all baseline models across nearly all tasks and datasets, achieving 46.0\%--75.9\% improvement in categorical metrics and 80.7\%--112.7\% in numerical metrics compared to industry-leading models like GPT-4o. This demonstrates that synthetic training data can effectively enable strong alignment with real-world time series.

Among the baselines, GPT-4o (Vision) performs best, suggesting vision-based MLLMs possess some capability to analyze shape characteristics of time series, though they remain limited by image resolution when interpreting details. Text-based methods struggle with the constraints of prompt length, while agent-based approaches performed below expectations (see Section~\ref{sec:agent_study} for detailed analysis).

{
\footnotesize
\begin{table}[!t]
    \centering
    \caption{Reasoning tasks. Inductive Reasoning is in the form of Q\&A, evaluated with RAGAS. Other tasks are MC or T/F questions, which are evaluated with accuracy.}
    \label{tab:reason}
    \begin{tabular}{cl|cccc|c}
        \toprule
        \textbf{Type} & \textbf{Model} & Induct. & Deduct. & Causal & MCQ2 & \textbf{Average} \\
        \midrule
        \multirow{4}{*}{Text} & GPT-4o-mini & 0.333 & 0.326 & 0.576 & 0.480 & 0.429 \\
        & GPT-4o & 0.336 & 0.628 & 0.685 & 0.470 & 0.530 \\
        & GPT-4-Turbo & 0.280 & 0.581 & 0.644 & 0.490 & 0.499 \\
        & QWen2.5-14B & 0.184 & 0.605 & 0.348 & 0.320 & 0.364 \\
        \midrule
        \multirow{2}{*}{Vision} & GPT-4o-mini & 0.323 & 0.442 & 0.495 & 0.480 & 0.435 \\
        & GPT-4o & 0.322 & 0.605 & 0.652 & 0.490 & 0.517 \\
        \midrule
        \multirow{2}{*}{Agent} & {GPT-4o-mini} & {0.219} & {0.357} & {0.692} & {0.340} & {0.402} \\
        & {GPT-4o} & {0.167} & {0.553} & {0.696} & {0.380} & {0.449} \\
        \midrule
        \multirow{1}{*}{TS} & \textbf{\modelname{}} & \textbf{0.518} & \textbf{0.744} & \textbf{0.804} & \textbf{0.600} & \textbf{0.667} \\
        \bottomrule
    \end{tabular}
    
\end{table}
}

ChatTS's advantages are particularly pronounced in multivariate tasks, where text-based models face challenges with excessively long prompts and vision-based models struggle to distinguish features across multiple time series plotted simultaneously. In contrast, ChatTS's context-aware time series encoding accurately analyzes referenced time series based on contextual information.

From the efficiency perspective, \modelname{}'s native multimodal encoding requires significantly fewer tokens to represent time series data, resulting in much lower costs compared with the baselines (see Table \ref{tab:align}). This shows both the effectiveness and efficiency of treating time series as a native modality.

\subsection{RQ2. Reasoning Tasks}
The comparison results of our model and the baseline models for Reasoning Tasks are shown in Table \ref{tab:reason}.
Reasoning tasks are typically more complex and better aligned with real-world application scenarios than alignment tasks.
It can be found that \modelname{} achieves consistent improvements over the baseline models across all reasoning tasks.
In the Inductive Reasoning task, \modelname{} achieved a 34.5\% improvement compared to the baseline models, indicating that \modelname{} can accurately associate time series attributes with their physical meanings in the real world. This demonstrates that the proposed attribute-based time series generation effectively enables the model to understand the patterns of the physical world reflected in time series. 
Moreover, \modelname{} also achieved notable improvements in other reasoning tasks, which indicates that even with only synthetic training data, the model can be equipped with good reasoning capabilities related to time series. This further demonstrates the effectiveness of the proposed attribute-based time series generation method and TSEvol.

\begin{figure*}[!htb]
\begin{subfigure}{\linewidth}
  \centering
  \includegraphics[width=0.95\linewidth]{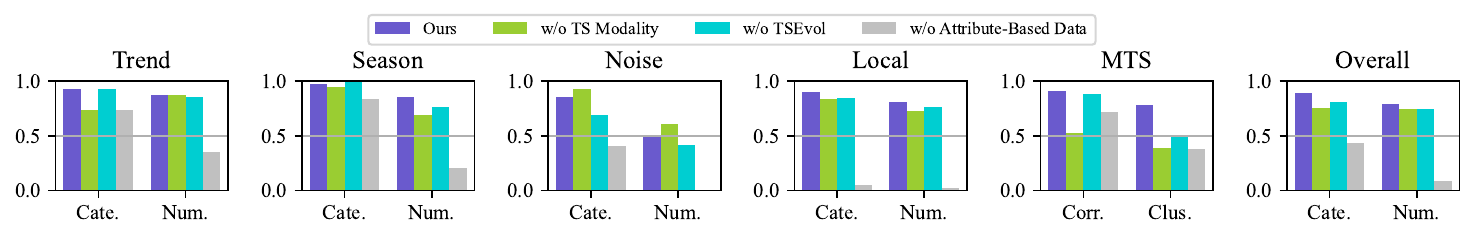}
  \caption{Dataset A}
  \label{fig:ablation_study_a}
\end{subfigure}\hfill
\begin{subfigure}{\linewidth}
  \centering
  \includegraphics[width=0.95\linewidth]{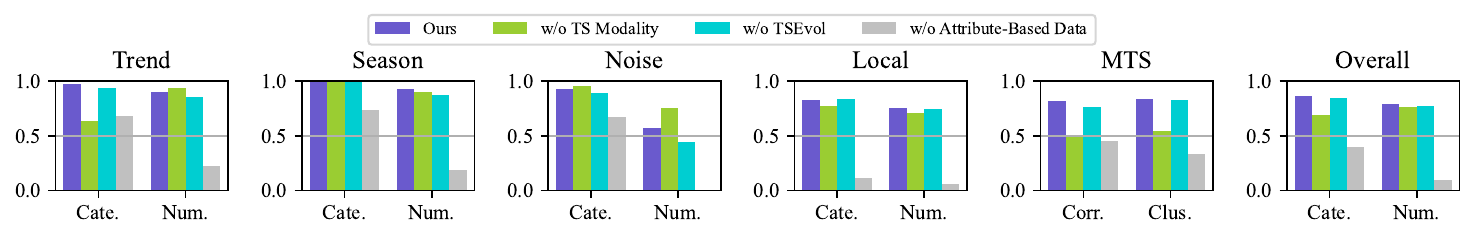}
  \setlength{\belowcaptionskip}{-10pt}
  \caption{Dataset B}
  \label{fig:ablation_study_b}
\end{subfigure}
\setlength{\belowcaptionskip}{-10pt}
\caption{Ablation studies on alignment tasks.}
\label{fig:ablation_study}
\end{figure*}

\begin{figure}[!htbp]
    \centering
    \includegraphics[width=0.85\linewidth]{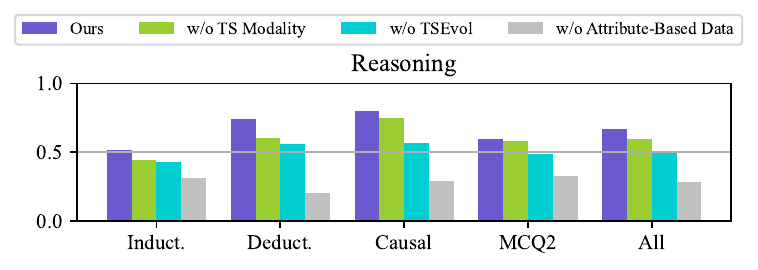}
    \setlength{\belowcaptionskip}{-10pt}
    \caption{Ablation studies on reasoning tasks.}
    \label{fig:ablation_study_reasoning}
\end{figure}

\subsection{RQ3. Studies of Synthetic Training Data}
To evaluate the effectiveness of attribute-based time series generation and TSEvol, we conducted ablation studies with two variants: (1) \textit{w/o Attribute-Based}, where all training datasets were replaced by GPT-generated datasets from~\cite{merrill2024language}, containing time series directly generated using GPT-produced Python code with corresponding GPT-generated Q\&As; (2) \textit{w/o TSEvol}, where SFT datasets were replaced with data directly generated using an LLM without the evolutionary approach, though with prompts designed to encourage diversity. Both variants included the instruct-following dataset to ensure fair comparison.

The evaluation results in Figures~\ref{fig:ablation_study} and~\ref{fig:ablation_study_reasoning} reveal that models trained on GPT-generated data performed significantly worse across alignment tasks, particularly for local fluctuation detection and numerical analysis. This suggests the attribute-based generation method better captures precise feature details and numerical values. Meanwhile, models trained with TSEvol demonstrated substantial improvements in reasoning capabilities and modest gains in alignment tasks, indicating that TSEvol effectively diversifies question formats and generates tailored Q\&As for different time series attributes, enhancing overall model performance.

\subsection{RQ4. Scaling of Training Dataset}
\label{sec:dataset_scaling}
\begin{figure}[!htbp]
\begin{subfigure}{0.27\linewidth}
  \centering
  \includegraphics[width=1.0\linewidth]{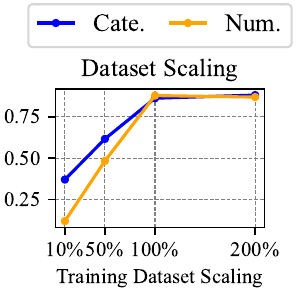}
  \caption{Train Data}
  \label{fig:dataset_scaling_study}
\end{subfigure}\hfill
\begin{subfigure}{0.72\linewidth}
  \centering
  \includegraphics[width=1.0\linewidth]{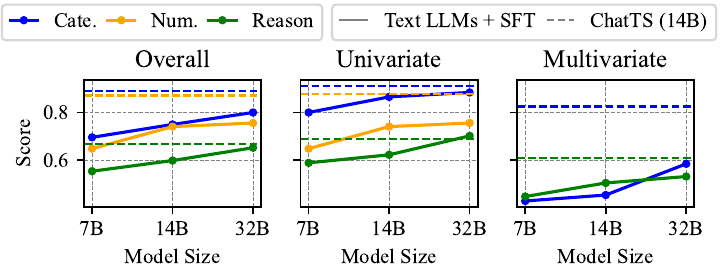}
  \caption{SFT on Text-Based LLMs}
  \label{fig:text_model_study}
\end{subfigure}\hfill
\setlength{\belowcaptionskip}{-10pt}
\caption{Scaling of training dataset and text-based LLMs.}
\label{fig:scaling_study}
\end{figure}
Figure~\ref{fig:dataset_scaling_study} illustrates the relationship between ChatTS performance and training data size. The results show that increasing the Phase 1 training dataset size from 10\% to 100\% of the current size significantly improves performance, but further expansion yields minimal gains. Thus, our chosen training set size is well-balanced, ensuring sufficient data for effective alignment while avoiding too much resource consumption during training.

\subsection{RQ5. Study of Time Series Modality}
To investigate the effectiveness of the time series multimodality in \modelname{}, we performed an ablation study based on a text-only version of \modelname{} (w/o TS Modality).
We remove the time series encoder in \modelname{} (\textit{i.e.} using the original QWen-2.5 model) and use the same training data with \modelname{} (the time series arrays are encoded into text) in model training.
The experimental results are shown in Figure \ref{fig:ablation_study} and Figure \ref{fig:ablation_study_reasoning}. 
Overall, the model using only the text modality performs significantly worse than the original \modelname{} model.
This indicates that encoding multimodal information is crucial for accurately capturing both shape and numerical information.
However, in certain sub-evaluation metrics (e.g., noise), the text-only model outperforms the multimodal \modelname{}, suggesting that text modality models still have strong capabilities for identifying small fluctuations.
In MTS tasks, the text-only model is nearly incapable of answering any questions. This implies that even with extensive multivariate training data, text-only LLMs still struggle to handle multivariate problems due to excessively long context lengths because of severe hallucinations and inaccurate responses.
Additionally, to compare the performance gains between text-based LLMs and TS-MLLMs, we fine-tuned various sizes of the Qwen2.5 series text-based LLMs using the text version of the ChatTS training dataset (as shown in Figure \ref{fig:text_model_study}). Experimental results indicate that even when fine-tuning the larger Qwen2.5-32B text model, the results still do not outperform those of \modelname{} (14B), which has native multimodal capabilities. This further validates the importance of native multimodal capabilities in \modelname{}, whether in the accuracy for MTS analysis or cost efficiency (see Table \ref{tab:align}).

\subsection{RQ6. Study of Agent-Based Methods}
\label{sec:agent_study}
\begin{figure}[!htbp]
\begin{subfigure}{1.0\linewidth}
  \centering
  \includegraphics[width=1.0\linewidth]{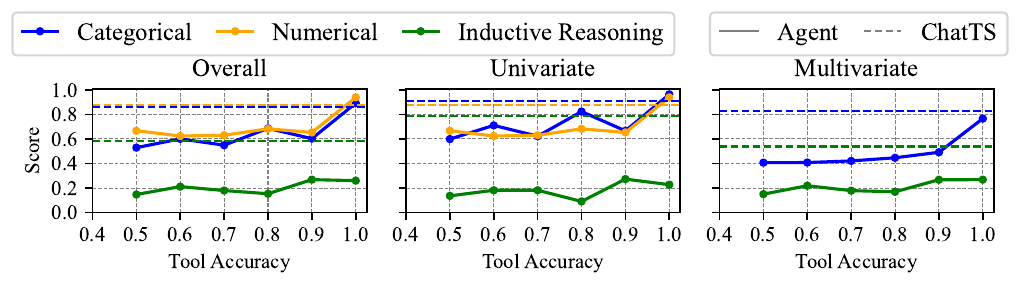}
  \caption{Agent with different tool accuracy.}
  \label{fig:agent_study_accuracy}
\end{subfigure}\hfill
\begin{subfigure}{0.56\linewidth}
  \centering
  \includegraphics[width=1.0\linewidth]{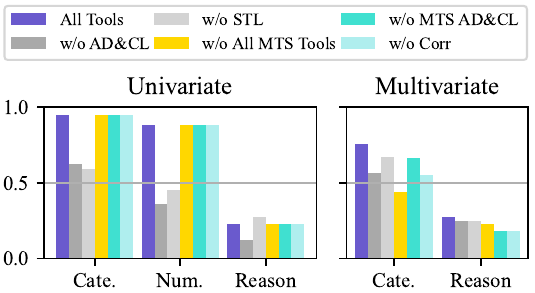}
    \caption{Agent w/o different tools.}
    \label{fig:agent_tool_study}
\end{subfigure}
\begin{subfigure}{0.43\linewidth}
  \centering
  \includegraphics[width=1.0\linewidth]{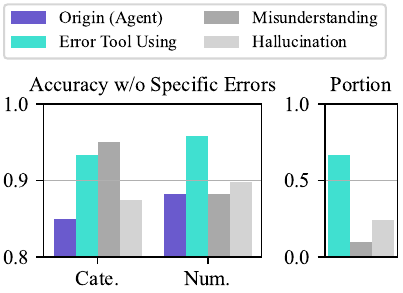}
  \caption{Study of error cases.}
  \label{fig:agent_error_study}
\end{subfigure}
\setlength{\belowcaptionskip}{-10pt}
\caption{Agent with ``Perfect Tools''. Even with perfect tools, Agent still makes errors (e.g., Error Tool Using)}
\label{fig:agent_study}
\end{figure}

Agent-based methods are widely applied but showed suboptimal performance in our evaluations (RQ1, RQ2) due to several main issues: (1) tool inaccuracy (2) error tool use, and (3) response formatting that caused parsing failures. To explore their performance upper bound, we conducted detailed analyses:

\begin{enumerate}[leftmargin=*]
    \item \textbf{Parsing Failures:} We exclude responses that failed to parse, ensuring all outputs were valid. 
    \item \textbf{Perfect Tools:} We design ``perfect tools'' with controlled accuracy through time series labels in the synthetic dataset (Figure \ref{fig:agent_study_accuracy}). The accuracy of the perfect tools can be strictly controlled.
    \item \textbf{Tool Ablation:} We perform ablation studies (Figure \ref{fig:agent_tool_study}) to evaluate the impact of individual tools on accuracy.
    \item \textbf{Error Analysis:} We categorize agent errors into three types: Error Tool Using, Misunderstanding, and Hallucination. We analyze their impact on performance (Figure \ref{fig:agent_error_study}).
\end{enumerate}

Our sensitivity analysis (Figure \ref{fig:agent_study_accuracy}) shows that agent performance is highly sensitive to tool accuracy, especially in the [0.9, 1.0] range. For categorical and numerical tasks, the agent with perfect tools slightly outperforms \modelname{} on UTS tasks but lags behind in MTS tasks. For agents, MTS tasks typically require more tool calls and reasoning, which places higher demands on LLMs' tool-using and summarization capabilities. In contrast, \modelname{} processes multiple time series natively, reducing complexity and improving accuracy for MTS tasks.
The tool ablation study (Figure \ref{fig:agent_tool_study}) shows that Agent depends heavily on both tool precision and completeness, particularly for MTS tasks. Even with all tools available, agents frequently fail to invoke the correct tool at the right time (e.g., using the classification tool rather than the anomaly detection tool to identify the position of a spike), limiting their effectiveness. Error analysis (Figure \ref{fig:agent_error_study}) reveals ``Error Tool Using'' as the largest source of errors. When these cases are excluded, agent accuracy exceeds 95\%, surpassing \modelname{}. This validates the correctness of the implementation of perfect tools and the model, which also shows their limitations: Agents may struggle with tool selection and reasoning.

In summary, while perfect tools improve agent performance, challenges such as tool selection errors, misunderstandings, and hallucinations persist, leaving agents less effective than ChatTS for complex time series tasks.

\section{Case Studies and Applications}
\label{sec:case_study}
\subsection{Case Studies on Real-World Data}
\begin{figure}[!htbp]
    \centering
    \includegraphics[width=1.0\linewidth]{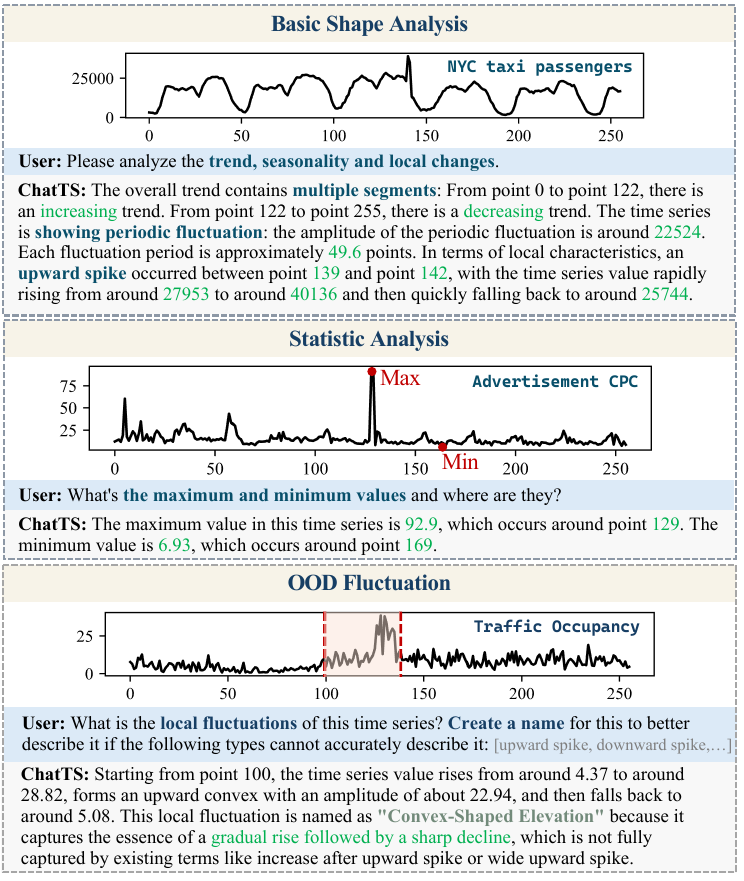}
    \caption{Case studies on real-world time series data.}
    \label{fig:case_study_real}
\end{figure}

To investigate the performance of \modelname{} on \textit{real-world} time series, we perform several case studies with challenging questions, the results are shown in Figure~\ref{fig:case_study_real}.

\subsubsection{Shape and Statistical Analysis}
The ``Basic Shape Analysis'' case demonstrates \modelname{}'s capability to analyze an NYC taxi passenger time series with complex periodic fluctuations and local anomalies. \modelname{} accurately identifies multiple trend segments, the periodicity, and the upward spike along with their amplitudes. This shows \modelname{}'s capability to capture both global patterns and localized features. In the ``Statistic Analysis'' case, \modelname{} analyzes advertisement CPC data with misleading scaling. Despite potential confusion in the minimum value, \modelname{} correctly identifies the max/min values and their positions. These cases show \modelname{}'s robustness in statistical analysis in complex real-world time series.

\subsubsection{OOD Fluctuation Recognition}
The ``OOD Fluctuation'' case presents \modelname{} with a traffic occupancy time series containing an OOD fluctuation pattern \textit{absent from its training data}. However, \modelname{} accurately describes it as a ``Convex-Shaped Elevation'', characterized by a gradual rise followed by a sharper decline along with the overall shape of a convex. This demonstrates \modelname{}'s inherent understanding of time series patterns themselves, rather than simply repeating representations from the training set. This indicates that \modelname{} has a certain capacity to generalize to real-world data despite being trained exclusively on synthetic data.

\subsection{Real-World Application: DB Operation}
\begin{figure}[!htbp]
    \centering
    \includegraphics[width=1.0\linewidth]{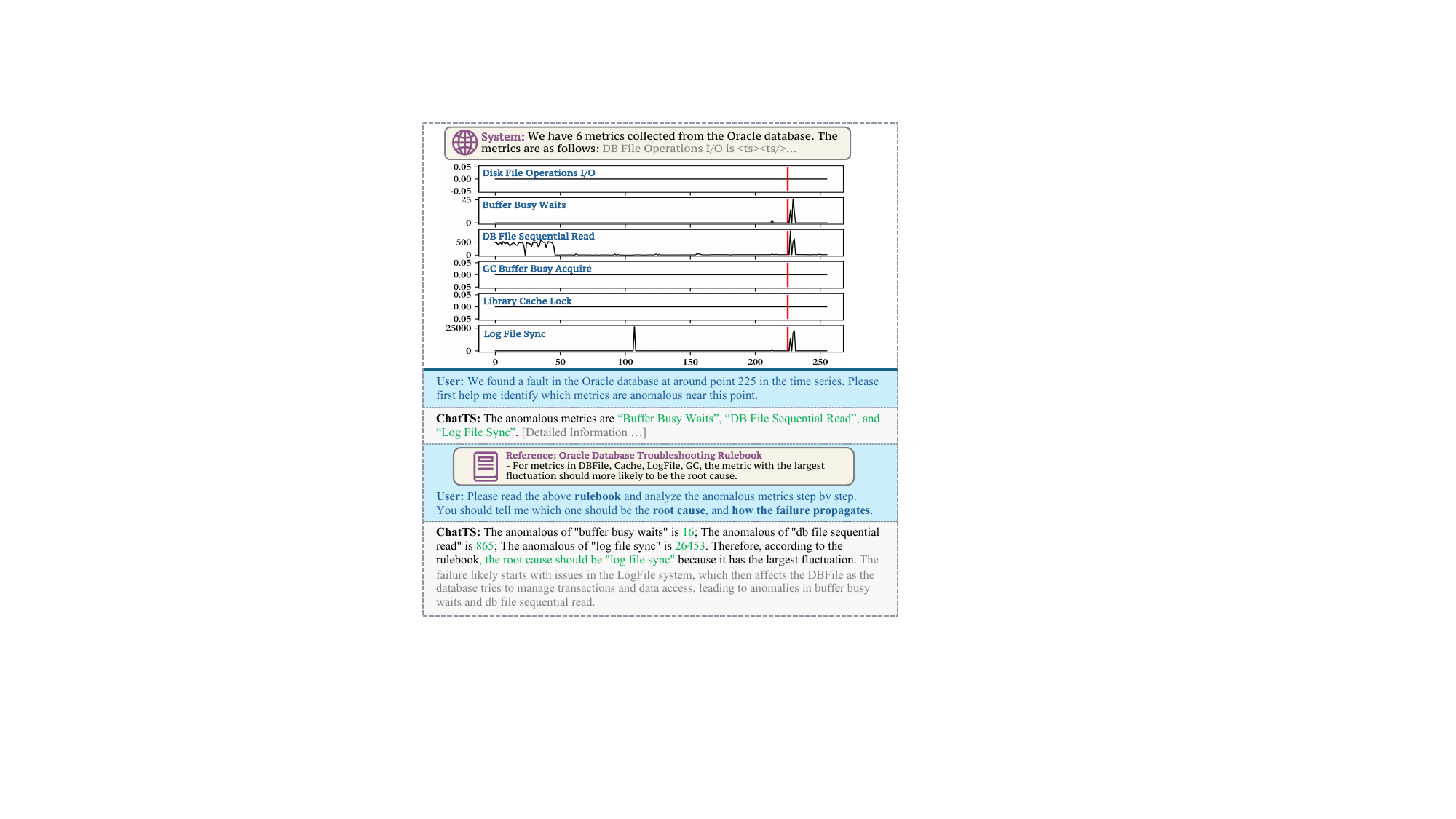}
    \caption{An application case of \modelname{} in a failure diagnosis with an Oracle database system.}
    \label{fig:case_study_aiops}
\end{figure}

To illustrate the performance of \modelname{} with its native time series multimodal capability in real-world applications, we present a typical Oracle DB operation application through an MTS-related multi-turn dialogue with \modelname{}.
In this case study, an Oracle DB operator has identified a recent anomaly and retrieves several time series metrics from the monitoring system, inputting them into \modelname{} for analysis (as shown in Figure \ref{fig:case_study_aiops})\footnote{DB Metrics can be input into \modelname{} through API. The system implementation details are out of the scope of this paper.}.
By querying \modelname{}, the operator obtains the names of all metrics with anomalies. Then, to accurately pinpoint the root cause, the operator provides \modelname{} with a textual document titled ``Oracle Database Troubleshooting Rulebook'' and requests \modelname{} to analyze the root cause and propagation of the system failure step-by-step, combining insights from the rulebook and the time series anomalies.
Notably, the rulebook is entirely in \textit{text} form, without a strictly structured format, which is helpful for the operators to share their expert experience effectively.
The responses of \modelname{} show that it can accurately identify anomalies and amplitudes in multivariate time series.
By leveraging ``the metric with the largest fluctuation'' in the rulebook, \modelname{} can further reason about the root cause and failure propagation path.
This further shows that \modelname{} can effectively utilize its \textit{alignment capability} to analyze time series and perform complex analysis in real-world applications with its robust \textit{reasoning ability}.

\subsection{Real-World Application: Detailed Analysis}
\begin{figure}[!htbp]
    \centering
    \includegraphics[width=1.0\linewidth]{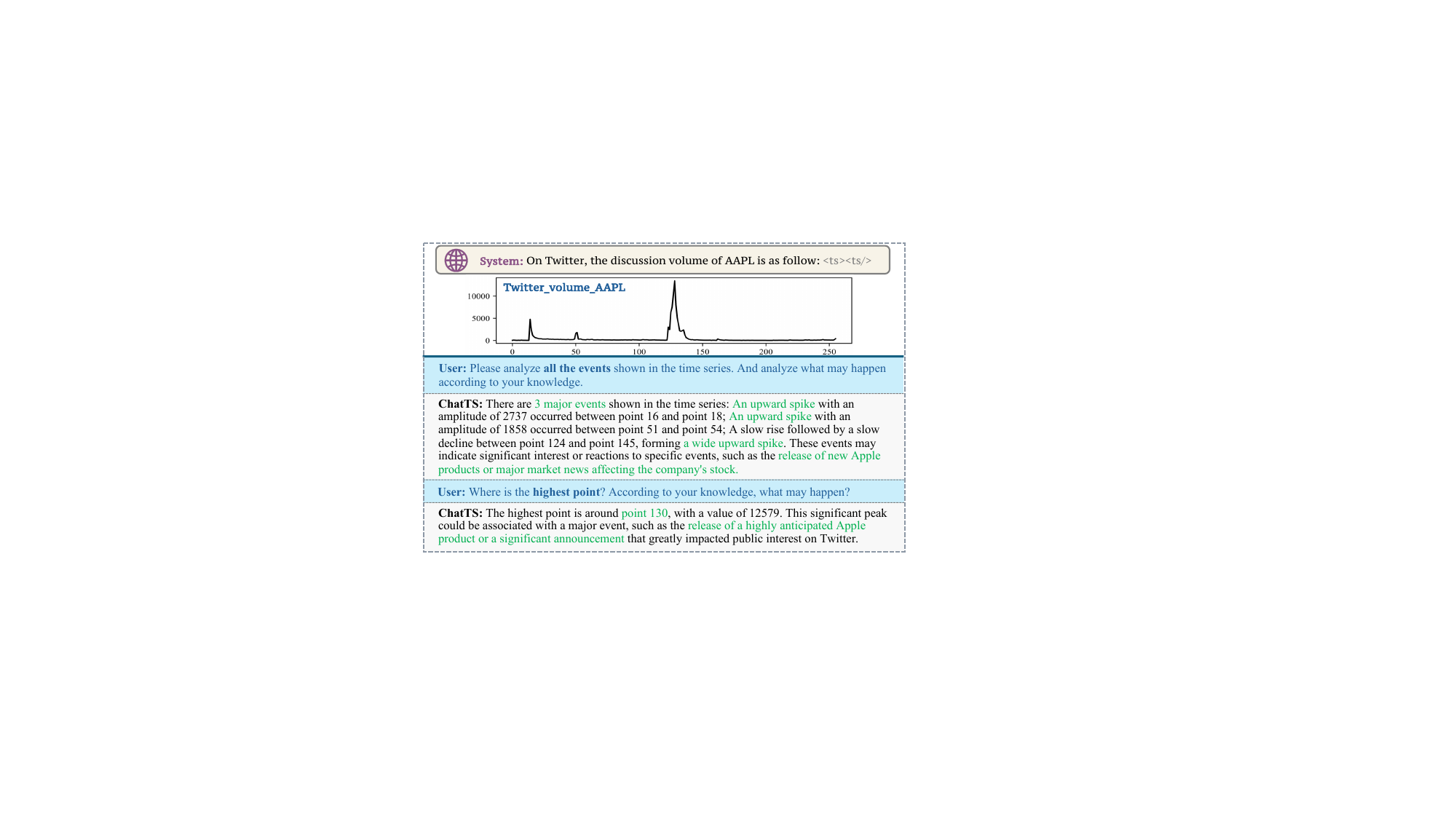}
    \caption{An \modelname{} application case in detailed time series.}
    \label{fig:case_study_aapl}
\end{figure}
Another typical application of \modelname{} is conducting a detailed analysis of time series features, combined with LLMs' knowledge and reasoning capabilities to perform simple reasoning and question answering.
In Figure \ref{fig:case_study_aapl}, we present a case study of time series analysis on the discussion intensity of AAPL-related topics on Twitter, using data from NAB~\cite{ahmad2017unsupervised}.
Notably, even without explicit instructions from the user to identify local fluctuations, \modelname{} can accurately infer the user's intent and determine the timestamps of all three ``hot events'' from the time series. Furthermore, \modelname{} can precisely identify the highest point and its position in the time series based on the numerical values of the local peaks and perform event analysis according to the physical meaning of the series. This demonstrates that \modelname{} can accurately recognize both shape and numerical characteristics of time series and perform reasoning and analysis based on vague user input.

\subsection{Baseline Comparison}
\begin{figure}[!htb]
\begin{subfigure}{\linewidth}
  \centering
  \includegraphics[width=1.0\linewidth]{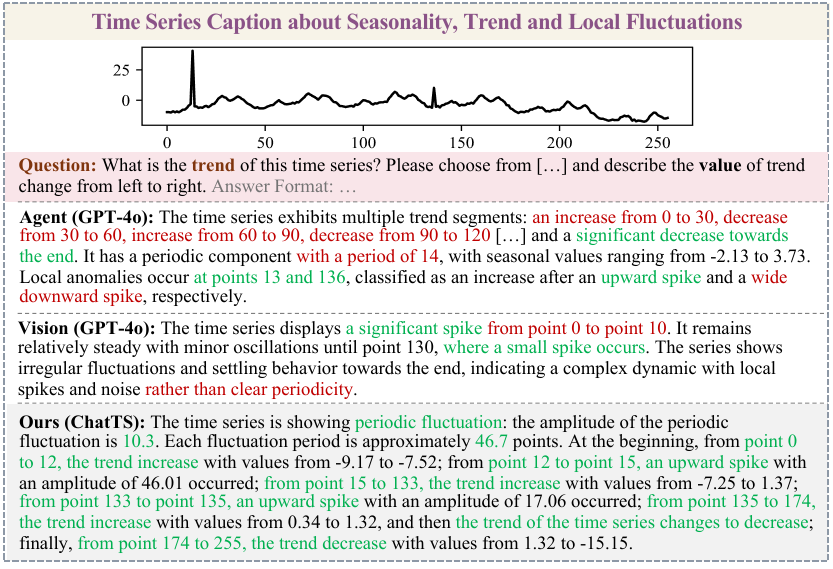}
  \caption{Case Study on Complex Time Series Caption}
  \label{fig:case_study_caption}
\end{subfigure}\hfill
\begin{subfigure}{\linewidth}
  \centering
  \includegraphics[width=1.0\linewidth]{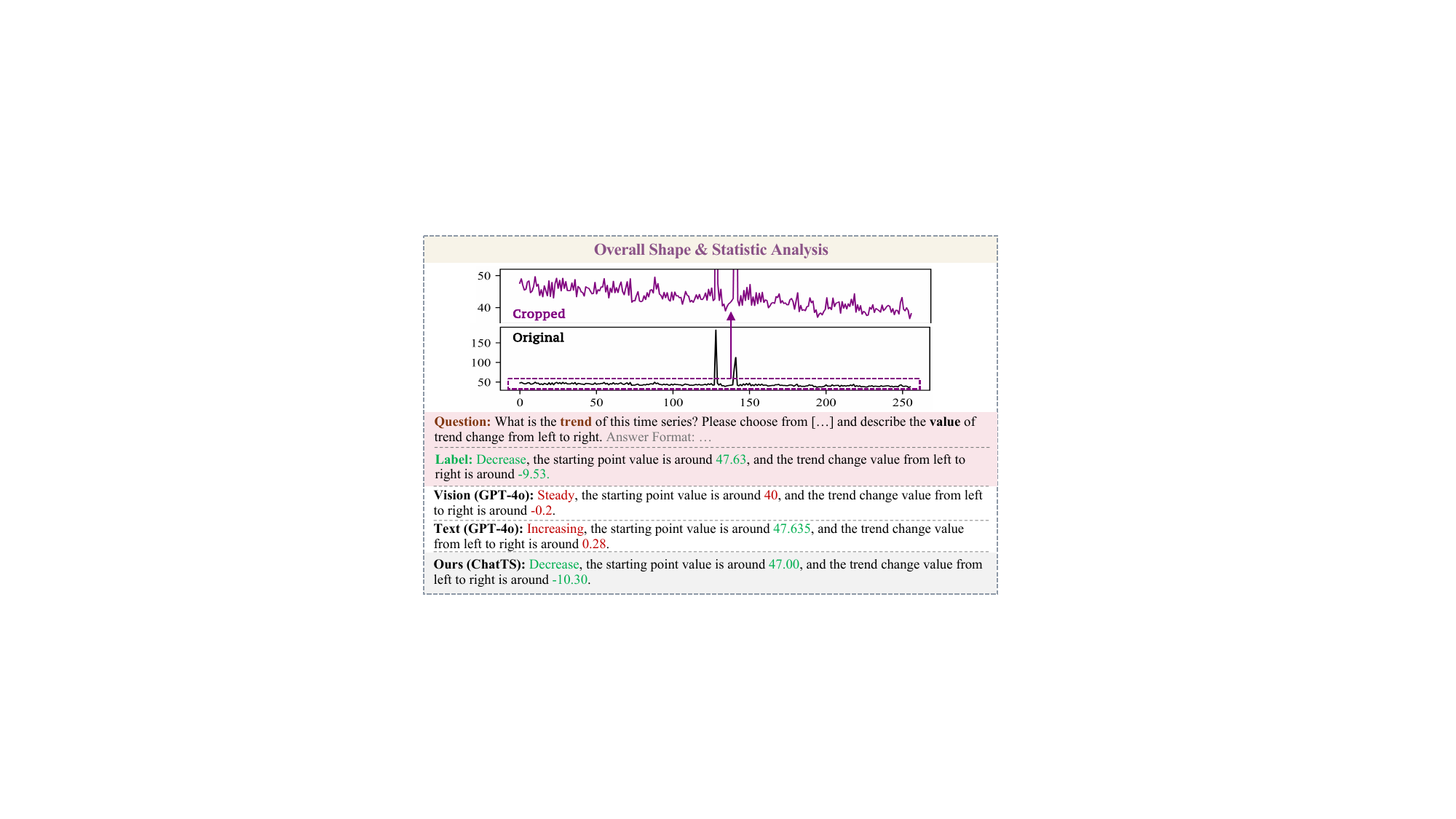}
  \caption{Case Study on Trend}
  \label{fig:case_study_2}
\end{subfigure}
\setlength{\belowcaptionskip}{-10pt}
\caption{Case studies on different types of LLMs in time series alignment tasks.}
\label{fig:case_study_align}
\end{figure}

\subsubsection{Seasonality, Trend and Fluctuations}
As shown in Figure~\ref{fig:case_study_caption}, due to tool inaccuracy, the Agent fails to identify the periodicity. This further led the Agent to misinterpret periodic patterns as different trend changes, resulting in errors in trend analysis. Moreover, the LLM does not realize the problem or attempt to correct it. Similarly, the Vision-based model also exhibited errors in analyzing local fluctuations and periodicity. In contrast, \modelname{}, with its time-series modality awareness, accurately captures the periodicity and trend transitions. This case shows a key limitation of agent-based tools: the precision of tools alone cannot overcome the cascading errors caused by the initial error of time series patterns.

\subsubsection{Detailed Trend Analysis}
Figure~\ref{fig:case_study_2} presents a misleading case where the original image suggests a steady trend. However, in the cropped plot that ignores the two spikes, there is a significant decreasing trend. Due to the subtle nature of trends displayed in the time series images, the Vision-based model incorrectly classified them as steady.
Similarly, text-based models, while identifying starting values, fail to identify the global shape of the time series.
In contrast, \modelname{} captures both the overall trend and numerical details accurately due to its native time series encoding capabilities.

\section{Related Work}
\textbf{Multimodal LLMs (MLLMs).}
MLLMs have developed rapidly in recent years and found extensive applications~\cite{zhang2024mm,yin2024survey}. 
A significant body of research integrates different types of data to achieve multimodal fusion, including images~\cite{liu2024visual,bai2023qwen,li2023blip}, videos~\cite{maaz2023video,zhang2023video,li2023videochat}, audio~\cite{chu2023qwen,rubenstein2023audiopalm}, and graphs~\cite{zhang2024graphtranslator,pan2024unifying}. These models have been applied across diverse domains, with image-based question answering and reasoning representing an important research direction. Many studies leverage vision-based LLMs for image reasoning tasks~\cite{luo2024mmevol,jiang2023cross}, fully utilizing the natural language understanding and reasoning capabilities of large language models. 
However, in the field of time series, despite the existence of numerous works (as discussed below) that combine time series data with LLMs, research on aligning LLMs with time-series modalities with time-series modalities remains limited. This limitation is primarily due to the scarcity of high-quality multimodal datasets that combine time series with textual information \cite{merrill2024language,chow2024towards,jin2024position}. As a result, the development of time series-specific MLLMs for question-answering and reasoning tasks has lagged behind other modalities.

\noindent \textbf{Time Series Question Answering (TSQA).}
With the rapid development of LLMs, TSQA systems have combined the reasoning capabilities of LLMs with time series analysis to enable more efficient cross-domain decision-making and complex task handling~\cite{jin2024position}. Time series question-answering systems have been explored in various fields, such as AIOps~\cite{dbot,rcagent}, IoT~\cite{xing2021deepsqa,gallo2023conversational}, healthcare~\cite{yu2023zero,oh2024ecg}, finance~\cite{maitre2020event,kurisinkel2024text2timeseries}, and traffic~\cite{da2024open,lai2023large}. However, these methods are often limited to agent-based~\cite{yao2022react} and retrieval-augmented generation (RAG)~\cite{lewis2020retrieval} approaches, lacking a comprehensive understanding of time series and sufficient reasoning capabilities. Although some recent studies~\cite{chow2024towards} have attempted to leverage temporal multimodal approaches for time series reasoning tasks, they typically rely on task-specific corpora. They are trained and evaluated on specific tasks (e.g., classification tasks or forecasting tasks), lacking multivariate analysis capabilities. Compared to the research on multimodal question answering in fields like images and videos, time series question answering still lacks robust multimodal alignment methods and evaluation frameworks~\cite{merrill2024language,cai2024timeseriesexam}. Therefore, in contrast to existing studies, this paper is the first to propose a comprehensive time series modality alignment and fine-tuning process, evaluated using multiple alignment and reasoning tasks.

\noindent \textbf{LLM + Time Series.}
In addition to the research above, many studies have combined LLMs with time series for various downstream tasks, leveraging the powerful capabilities of LLMs~\cite{jin2023time,gruver2024large,su2024large,chang2023llm4ts,liu2024time,cao2023tempo,zhou2023one,cai2023jolt}. However, while these models are using LLMs as backbones, they are designed for specific downstream tasks and lack language alignment capabilities, making them unsuitable for question answering and reasoning applications. Moreover, some studies employ vision-based multimodal LLMs for time series prediction~\cite{chen2024visionts} and anomaly detection~\cite{zhuang2024see}. This approach aligns with the vision-based LLM methods discussed in this paper but is significantly constrained in its ability to analyze time series.


\section{Limitation and Future Work}
Due to the limited existing research on time series understanding and reasoning, although \modelname{} has explored an effective approach, we believe it still has a number of limitations.
First, while our experiments demonstrate that synthetic data can achieve satisfactory alignment and reasoning performance, we believe that real-world data is essential for further enhancing the capabilities of TS-MLLMs. We hope more relevant datasets will emerge in the future. Second, although we found that a simple MLP encoder performs well due to the relatively simple structure of time series data, exploring more effective methods for multimodal encoding and integration remains a valuable research direction. Third, despite labeling hundreds of real-world time series and using 14 evaluation metrics for evaluation, we believe that this is still insufficient for a comprehensive evaluation of TS-MLLMs. More labeled real-world data is needed for a more comprehensive evaluation. Finally, while this work focuses on \textit{understanding tasks} like language alignment and reasoning, MLLM-based time series \textit{generation} is also worth exploring. Thus, developing a multimodal model that can generate time series based on textual input is an important area for future research.

\section{Conclusion}
Understanding and reasoning are important for real-world time series applications, but research is limited due to the lack of time series-text data.
In this paper, we propose \modelname{}, the first TS-MLLM with multivariate time series as input for complex time series QA and reasoning, which is fine-tuned on synthetic data.
We introduce an attribute-based time series generation method, which not only generates diverse time series but also provides complete and precise attribute descriptions.
Building on this, we further propose TSEvol, which leverages rich attribute combinations from the attribute pool and Evol-Instruct to generate diverse and accurate QAs, enhancing the model's capabilities in complex question answering and reasoning.
To comprehensively evaluate the capabilities of our model, we collect datasets that include real-world time series data, covering the evaluation of both alignment tasks and reasoning tasks.
Evaluation results show that our model achieves significant improvements, outperforming baselines by 46.0\% in alignment tasks and 25.8\% in reasoning tasks. These findings demonstrate the effectiveness of our approach in bridging the gap between time series data and natural language understanding.
We have open-sourced the source code, trained model weights, and the evaluation datasets for reproduction and future research: \url{https://github.com/NetManAIOps/ChatTS}.

\bibliographystyle{ACM-Reference-Format}
\bibliography{ref}


\begin{thebibliography}{69}


\ifx \showCODEN    \undefined \def \showCODEN     #1{\unskip}     \fi
\ifx \showDOI      \undefined \def \showDOI       #1{#1}\fi
\ifx \showISBNx    \undefined \def \showISBNx     #1{\unskip}     \fi
\ifx \showISBNxiii \undefined \def \showISBNxiii  #1{\unskip}     \fi
\ifx \showISSN     \undefined \def \showISSN      #1{\unskip}     \fi
\ifx \showLCCN     \undefined \def \showLCCN      #1{\unskip}     \fi
\ifx \shownote     \undefined \def \shownote      #1{#1}          \fi
\ifx \showarticletitle \undefined \def \showarticletitle #1{#1}   \fi
\ifx \showURL      \undefined \def \showURL       {\relax}        \fi
\providecommand\bibfield[2]{#2}
\providecommand\bibinfo[2]{#2}
\providecommand\natexlab[1]{#1}
\providecommand\showeprint[2][]{arXiv:#2}

\bibitem[\protect\citeauthoryear{??}{adt}{2019}]%
        {adtk}
 \bibinfo{year}{2019}\natexlab{}.
\newblock \bibinfo{title}{Anomaly Detection Toolkit.}
\newblock
\newblock
\urldef\tempurl%
\url{https://github.com/arundo/adtk}
\showURL{%
\tempurl}


\bibitem[\protect\citeauthoryear{??}{dee}{2020}]%
        {deepspeed}
 \bibinfo{year}{2020}\natexlab{}.
\newblock \bibinfo{title}{DeepSpeed.}
\newblock
\newblock
\urldef\tempurl%
\url{https://www.deepspeed.ai/}
\showURL{%
\tempurl}


\bibitem[\protect\citeauthoryear{??}{wea}{2023}]%
        {weatherdataset}
 \bibinfo{year}{2023}\natexlab{}.
\newblock \bibinfo{title}{Weather Dataset.}
\newblock
\newblock
\urldef\tempurl%
\url{https://www.bgc-jena.mpg.de/wetter/}
\showURL{%
\tempurl}


\bibitem[\protect\citeauthoryear{??}{mcq}{2024}]%
        {mcq2dataset}
 \bibinfo{year}{2024}\natexlab{}.
\newblock \bibinfo{title}{MCQ2 Dataset.}
\newblock
\newblock
\urldef\tempurl%
\url{https://github.com/behavioral-data/TSandLanguage}
\showURL{%
\tempurl}


\bibitem[\protect\citeauthoryear{??}{gpt}{2024}]%
        {gpt4o}
 \bibinfo{year}{2024}\natexlab{}.
\newblock \bibinfo{title}{OpenAI GPT-4o.}
\newblock
\newblock
\urldef\tempurl%
\url{https://openai.com/index/hello-gpt-4o/}
\showURL{%
\tempurl}


\bibitem[\protect\citeauthoryear{??}{qwe}{2024}]%
        {qwen14b}
 \bibinfo{year}{2024}\natexlab{}.
\newblock \bibinfo{title}{Qwen2.5-14B-Instruct Model.}
\newblock
\newblock
\urldef\tempurl%
\url{https://huggingface.co/Qwen/Qwen2.5-14B-Instruct}
\showURL{%
\tempurl}


\bibitem[\protect\citeauthoryear{Ahmad, Lavin, Purdy, and Agha}{Ahmad et~al\mbox{.}}{2017}]%
        {ahmad2017unsupervised}
\bibfield{author}{\bibinfo{person}{Subutai Ahmad}, \bibinfo{person}{Alexander Lavin}, \bibinfo{person}{Scott Purdy}, {and} \bibinfo{person}{Zuha Agha}.} \bibinfo{year}{2017}\natexlab{}.
\newblock \showarticletitle{Unsupervised real-time anomaly detection for streaming data}.
\newblock \bibinfo{journal}{\emph{Neurocomputing}}  \bibinfo{volume}{262} (\bibinfo{year}{2017}), \bibinfo{pages}{134--147}.
\newblock


\bibitem[\protect\citeauthoryear{Alnegheimish, Nguyen, Berti-Equille, and Veeramachaneni}{Alnegheimish et~al\mbox{.}}{2024}]%
        {alnegheimish2024large}
\bibfield{author}{\bibinfo{person}{Sarah Alnegheimish}, \bibinfo{person}{Linh Nguyen}, \bibinfo{person}{Laure Berti-Equille}, {and} \bibinfo{person}{Kalyan Veeramachaneni}.} \bibinfo{year}{2024}\natexlab{}.
\newblock \showarticletitle{Large language models can be zero-shot anomaly detectors for time series?}
\newblock \bibinfo{journal}{\emph{arXiv preprint arXiv:2405.14755}} (\bibinfo{year}{2024}).
\newblock


\bibitem[\protect\citeauthoryear{Bai, Bai, Yang, Wang, Tan, Wang, Lin, Zhou, and Zhou}{Bai et~al\mbox{.}}{2023}]%
        {bai2023qwen}
\bibfield{author}{\bibinfo{person}{Jinze Bai}, \bibinfo{person}{Shuai Bai}, \bibinfo{person}{Shusheng Yang}, \bibinfo{person}{Shijie Wang}, \bibinfo{person}{Sinan Tan}, \bibinfo{person}{Peng Wang}, \bibinfo{person}{Junyang Lin}, \bibinfo{person}{Chang Zhou}, {and} \bibinfo{person}{Jingren Zhou}.} \bibinfo{year}{2023}\natexlab{}.
\newblock \showarticletitle{Qwen-vl: A frontier large vision-language model with versatile abilities}.
\newblock \bibinfo{journal}{\emph{arXiv preprint arXiv:2308.12966}} (\bibinfo{year}{2023}).
\newblock


\bibitem[\protect\citeauthoryear{Cai, Choudhry, Goswami, and Dubrawski}{Cai et~al\mbox{.}}{2024}]%
        {cai2024timeseriesexam}
\bibfield{author}{\bibinfo{person}{Yifu Cai}, \bibinfo{person}{Arjun Choudhry}, \bibinfo{person}{Mononito Goswami}, {and} \bibinfo{person}{Artur Dubrawski}.} \bibinfo{year}{2024}\natexlab{}.
\newblock \showarticletitle{TimeSeriesExam: A time series understanding exam}.
\newblock \bibinfo{journal}{\emph{arXiv preprint arXiv:2410.14752}} (\bibinfo{year}{2024}).
\newblock


\bibitem[\protect\citeauthoryear{Cai, Goswami, Choudhry, Srinivasan, and Dubrawski}{Cai et~al\mbox{.}}{2023}]%
        {cai2023jolt}
\bibfield{author}{\bibinfo{person}{Yifu Cai}, \bibinfo{person}{Mononito Goswami}, \bibinfo{person}{Arjun Choudhry}, \bibinfo{person}{Arvind Srinivasan}, {and} \bibinfo{person}{Artur Dubrawski}.} \bibinfo{year}{2023}\natexlab{}.
\newblock \showarticletitle{Jolt: Jointly learned representations of language and time-series}. In \bibinfo{booktitle}{\emph{Deep Generative Models for Health Workshop NeurIPS 2023}}.
\newblock


\bibitem[\protect\citeauthoryear{Cao, Jia, Arik, Pfister, Zheng, Ye, and Liu}{Cao et~al\mbox{.}}{2023}]%
        {cao2023tempo}
\bibfield{author}{\bibinfo{person}{Defu Cao}, \bibinfo{person}{Furong Jia}, \bibinfo{person}{Sercan~O Arik}, \bibinfo{person}{Tomas Pfister}, \bibinfo{person}{Yixiang Zheng}, \bibinfo{person}{Wen Ye}, {and} \bibinfo{person}{Yan Liu}.} \bibinfo{year}{2023}\natexlab{}.
\newblock \showarticletitle{Tempo: Prompt-based generative pre-trained transformer for time series forecasting}.
\newblock \bibinfo{journal}{\emph{arXiv preprint arXiv:2310.04948}} (\bibinfo{year}{2023}).
\newblock


\bibitem[\protect\citeauthoryear{Chang, Peng, and Chen}{Chang et~al\mbox{.}}{2023}]%
        {chang2023llm4ts}
\bibfield{author}{\bibinfo{person}{Ching Chang}, \bibinfo{person}{Wen-Chih Peng}, {and} \bibinfo{person}{Tien-Fu Chen}.} \bibinfo{year}{2023}\natexlab{}.
\newblock \showarticletitle{Llm4ts: Two-stage fine-tuning for time-series forecasting with pre-trained llms}.
\newblock \bibinfo{journal}{\emph{arXiv preprint arXiv:2308.08469}} (\bibinfo{year}{2023}).
\newblock


\bibitem[\protect\citeauthoryear{Chen, Shen, Li, Wang, Sun, and Liu}{Chen et~al\mbox{.}}{2024}]%
        {chen2024visionts}
\bibfield{author}{\bibinfo{person}{Mouxiang Chen}, \bibinfo{person}{Lefei Shen}, \bibinfo{person}{Zhuo Li}, \bibinfo{person}{Xiaoyun~Joy Wang}, \bibinfo{person}{Jianling Sun}, {and} \bibinfo{person}{Chenghao Liu}.} \bibinfo{year}{2024}\natexlab{}.
\newblock \showarticletitle{VisionTS: Visual Masked Autoencoders Are Free-Lunch Zero-Shot Time Series Forecasters}.
\newblock \bibinfo{journal}{\emph{arXiv preprint arXiv:2408.17253}} (\bibinfo{year}{2024}).
\newblock


\bibitem[\protect\citeauthoryear{Chow, Gardiner, Hallgr{\'\i}msson, Xu, and Ren}{Chow et~al\mbox{.}}{2024}]%
        {chow2024towards}
\bibfield{author}{\bibinfo{person}{Winnie Chow}, \bibinfo{person}{Lauren Gardiner}, \bibinfo{person}{Haraldur~T Hallgr{\'\i}msson}, \bibinfo{person}{Maxwell~A Xu}, {and} \bibinfo{person}{Shirley~You Ren}.} \bibinfo{year}{2024}\natexlab{}.
\newblock \showarticletitle{Towards time series reasoning with llms}.
\newblock \bibinfo{journal}{\emph{arXiv preprint arXiv:2409.11376}} (\bibinfo{year}{2024}).
\newblock


\bibitem[\protect\citeauthoryear{Chu, Xu, Zhou, Yang, Zhang, Yan, Zhou, and Zhou}{Chu et~al\mbox{.}}{2023}]%
        {chu2023qwen}
\bibfield{author}{\bibinfo{person}{Yunfei Chu}, \bibinfo{person}{Jin Xu}, \bibinfo{person}{Xiaohuan Zhou}, \bibinfo{person}{Qian Yang}, \bibinfo{person}{Shiliang Zhang}, \bibinfo{person}{Zhijie Yan}, \bibinfo{person}{Chang Zhou}, {and} \bibinfo{person}{Jingren Zhou}.} \bibinfo{year}{2023}\natexlab{}.
\newblock \showarticletitle{Qwen-audio: Advancing universal audio understanding via unified large-scale audio-language models}.
\newblock \bibinfo{journal}{\emph{arXiv preprint arXiv:2311.07919}} (\bibinfo{year}{2023}).
\newblock


\bibitem[\protect\citeauthoryear{Da, Liou, Chen, Zhou, Luo, Yang, and Wei}{Da et~al\mbox{.}}{2024}]%
        {da2024open}
\bibfield{author}{\bibinfo{person}{Longchao Da}, \bibinfo{person}{Kuanru Liou}, \bibinfo{person}{Tiejin Chen}, \bibinfo{person}{Xuesong Zhou}, \bibinfo{person}{Xiangyong Luo}, \bibinfo{person}{Yezhou Yang}, {and} \bibinfo{person}{Hua Wei}.} \bibinfo{year}{2024}\natexlab{}.
\newblock \showarticletitle{Open-ti: Open traffic intelligence with augmented language model}.
\newblock \bibinfo{journal}{\emph{International Journal of Machine Learning and Cybernetics}} (\bibinfo{year}{2024}), \bibinfo{pages}{1--26}.
\newblock


\bibitem[\protect\citeauthoryear{Dempster, Petitjean, and Webb}{Dempster et~al\mbox{.}}{2020}]%
        {dempster2020rocket}
\bibfield{author}{\bibinfo{person}{Angus Dempster}, \bibinfo{person}{Fran{\c{c}}ois Petitjean}, {and} \bibinfo{person}{Geoffrey~I Webb}.} \bibinfo{year}{2020}\natexlab{}.
\newblock \showarticletitle{ROCKET: exceptionally fast and accurate time series classification using random convolutional kernels}.
\newblock \bibinfo{journal}{\emph{Data Mining and Knowledge Discovery}} \bibinfo{volume}{34}, \bibinfo{number}{5} (\bibinfo{year}{2020}), \bibinfo{pages}{1454--1495}.
\newblock


\bibitem[\protect\citeauthoryear{Es, James, Espinosa-Anke, and Schockaert}{Es et~al\mbox{.}}{2023}]%
        {es2023ragas}
\bibfield{author}{\bibinfo{person}{Shahul Es}, \bibinfo{person}{Jithin James}, \bibinfo{person}{Luis Espinosa-Anke}, {and} \bibinfo{person}{Steven Schockaert}.} \bibinfo{year}{2023}\natexlab{}.
\newblock \showarticletitle{Ragas: Automated evaluation of retrieval augmented generation}.
\newblock \bibinfo{journal}{\emph{arXiv preprint arXiv:2309.15217}} (\bibinfo{year}{2023}).
\newblock


\bibitem[\protect\citeauthoryear{Fons, Kaur, Palande, Zeng, Balch, Veloso, and Vyetrenko}{Fons et~al\mbox{.}}{2024}]%
        {fons2024evaluating}
\bibfield{author}{\bibinfo{person}{Elizabeth Fons}, \bibinfo{person}{Rachneet Kaur}, \bibinfo{person}{Soham Palande}, \bibinfo{person}{Zhen Zeng}, \bibinfo{person}{Tucker Balch}, \bibinfo{person}{Manuela Veloso}, {and} \bibinfo{person}{Svitlana Vyetrenko}.} \bibinfo{year}{2024}\natexlab{}.
\newblock \showarticletitle{Evaluating Large Language Models on Time Series Feature Understanding: A Comprehensive Taxonomy and Benchmark}.
\newblock \bibinfo{journal}{\emph{arXiv preprint arXiv:2404.16563}} (\bibinfo{year}{2024}).
\newblock


\bibitem[\protect\citeauthoryear{Fu, Chen, Zhang, Yang, Ma, and Yang}{Fu et~al\mbox{.}}{2024}]%
        {fu2024synthetic}
\bibfield{author}{\bibinfo{person}{Fanzhe Fu}, \bibinfo{person}{Junru Chen}, \bibinfo{person}{Jing Zhang}, \bibinfo{person}{Carl Yang}, \bibinfo{person}{Lvbin Ma}, {and} \bibinfo{person}{Yang Yang}.} \bibinfo{year}{2024}\natexlab{}.
\newblock \showarticletitle{Are Synthetic Time-series Data Really not as Good as Real Data?}
\newblock \bibinfo{journal}{\emph{arXiv preprint arXiv:2402.00607}} (\bibinfo{year}{2024}).
\newblock


\bibitem[\protect\citeauthoryear{Gallo, Paterno, and Malizia}{Gallo et~al\mbox{.}}{2023}]%
        {gallo2023conversational}
\bibfield{author}{\bibinfo{person}{Simone Gallo}, \bibinfo{person}{Fabio Paterno}, {and} \bibinfo{person}{Alessio Malizia}.} \bibinfo{year}{2023}\natexlab{}.
\newblock \showarticletitle{Conversational interfaces in iot ecosystems: where we are, what is still missing}. In \bibinfo{booktitle}{\emph{Proceedings of the 22nd International Conference on Mobile and Ubiquitous Multimedia}}. \bibinfo{pages}{279--293}.
\newblock


\bibitem[\protect\citeauthoryear{Gruver, Finzi, Qiu, and Wilson}{Gruver et~al\mbox{.}}{2024}]%
        {gruver2024large}
\bibfield{author}{\bibinfo{person}{Nate Gruver}, \bibinfo{person}{Marc Finzi}, \bibinfo{person}{Shikai Qiu}, {and} \bibinfo{person}{Andrew~G Wilson}.} \bibinfo{year}{2024}\natexlab{}.
\newblock \showarticletitle{Large language models are zero-shot time series forecasters}.
\newblock \bibinfo{journal}{\emph{Advances in Neural Information Processing Systems}}  \bibinfo{volume}{36} (\bibinfo{year}{2024}).
\newblock


\bibitem[\protect\citeauthoryear{He, Li, Tan, Wu, and Li}{He et~al\mbox{.}}{2023}]%
        {DBLP:journals/pvldb/HeLTWL23}
\bibfield{author}{\bibinfo{person}{Xiao He}, \bibinfo{person}{Ye Li}, \bibinfo{person}{Jian Tan}, \bibinfo{person}{Bin Wu}, {and} \bibinfo{person}{Feifei Li}.} \bibinfo{year}{2023}\natexlab{}.
\newblock \showarticletitle{OneShotSTL: One-Shot Seasonal-Trend Decomposition For Online Time Series Anomaly Detection And Forecasting}.
\newblock \bibinfo{journal}{\emph{Proc. {VLDB} Endow.}} \bibinfo{volume}{16}, \bibinfo{number}{6} (\bibinfo{year}{2023}), \bibinfo{pages}{1399--1412}.
\newblock
\urldef\tempurl%
\url{https://doi.org/10.14778/3583140.3583155}
\showDOI{\tempurl}


\bibitem[\protect\citeauthoryear{Jiang and Ye}{Jiang and Ye}{2023}]%
        {jiang2023cross}
\bibfield{author}{\bibinfo{person}{Ding Jiang} {and} \bibinfo{person}{Mang Ye}.} \bibinfo{year}{2023}\natexlab{}.
\newblock \showarticletitle{Cross-modal implicit relation reasoning and aligning for text-to-image person retrieval}. In \bibinfo{booktitle}{\emph{Proceedings of the IEEE/CVF Conference on Computer Vision and Pattern Recognition}}. \bibinfo{pages}{2787--2797}.
\newblock


\bibitem[\protect\citeauthoryear{Jin, Wang, Ma, Chu, Zhang, Shi, Chen, Liang, Li, Pan, et~al\mbox{.}}{Jin et~al\mbox{.}}{2023}]%
        {jin2023time}
\bibfield{author}{\bibinfo{person}{Ming Jin}, \bibinfo{person}{Shiyu Wang}, \bibinfo{person}{Lintao Ma}, \bibinfo{person}{Zhixuan Chu}, \bibinfo{person}{James~Y Zhang}, \bibinfo{person}{Xiaoming Shi}, \bibinfo{person}{Pin-Yu Chen}, \bibinfo{person}{Yuxuan Liang}, \bibinfo{person}{Yuan-Fang Li}, \bibinfo{person}{Shirui Pan}, {et~al\mbox{.}}} \bibinfo{year}{2023}\natexlab{}.
\newblock \showarticletitle{Time-llm: Time series forecasting by reprogramming large language models}.
\newblock \bibinfo{journal}{\emph{arXiv preprint arXiv:2310.01728}} (\bibinfo{year}{2023}).
\newblock


\bibitem[\protect\citeauthoryear{Jin, Zhang, Chen, Zhang, Liang, Yang, Wang, Pan, and Wen}{Jin et~al\mbox{.}}{2024}]%
        {jin2024position}
\bibfield{author}{\bibinfo{person}{Ming Jin}, \bibinfo{person}{Yifan Zhang}, \bibinfo{person}{Wei Chen}, \bibinfo{person}{Kexin Zhang}, \bibinfo{person}{Yuxuan Liang}, \bibinfo{person}{Bin Yang}, \bibinfo{person}{Jindong Wang}, \bibinfo{person}{Shirui Pan}, {and} \bibinfo{person}{Qingsong Wen}.} \bibinfo{year}{2024}\natexlab{}.
\newblock \showarticletitle{Position paper: What can large language models tell us about time series analysis}.
\newblock \bibinfo{journal}{\emph{arXiv preprint arXiv:2402.02713}} (\bibinfo{year}{2024}).
\newblock


\bibitem[\protect\citeauthoryear{Kurisinkel, Mishra, and Zhang}{Kurisinkel et~al\mbox{.}}{2024}]%
        {kurisinkel2024text2timeseries}
\bibfield{author}{\bibinfo{person}{Litton~Jose Kurisinkel}, \bibinfo{person}{Pruthwik Mishra}, {and} \bibinfo{person}{Yue Zhang}.} \bibinfo{year}{2024}\natexlab{}.
\newblock \showarticletitle{Text2timeseries: Enhancing financial forecasting through time series prediction updates with event-driven insights from large language models}.
\newblock \bibinfo{journal}{\emph{arXiv preprint arXiv:2407.03689}} (\bibinfo{year}{2024}).
\newblock


\bibitem[\protect\citeauthoryear{Lai, Xu, Zhang, Liu, and Xiong}{Lai et~al\mbox{.}}{2023}]%
        {lai2023large}
\bibfield{author}{\bibinfo{person}{Siqi Lai}, \bibinfo{person}{Zhao Xu}, \bibinfo{person}{Weijia Zhang}, \bibinfo{person}{Hao Liu}, {and} \bibinfo{person}{Hui Xiong}.} \bibinfo{year}{2023}\natexlab{}.
\newblock \showarticletitle{Large language models as traffic signal control agents: Capacity and opportunity}.
\newblock \bibinfo{journal}{\emph{arXiv preprint arXiv:2312.16044}} (\bibinfo{year}{2023}).
\newblock


\bibitem[\protect\citeauthoryear{Lewis, Perez, Piktus, Petroni, Karpukhin, Goyal, K{\"u}ttler, Lewis, Yih, Rockt{\"a}schel, et~al\mbox{.}}{Lewis et~al\mbox{.}}{2020}]%
        {lewis2020retrieval}
\bibfield{author}{\bibinfo{person}{Patrick Lewis}, \bibinfo{person}{Ethan Perez}, \bibinfo{person}{Aleksandra Piktus}, \bibinfo{person}{Fabio Petroni}, \bibinfo{person}{Vladimir Karpukhin}, \bibinfo{person}{Naman Goyal}, \bibinfo{person}{Heinrich K{\"u}ttler}, \bibinfo{person}{Mike Lewis}, \bibinfo{person}{Wen-tau Yih}, \bibinfo{person}{Tim Rockt{\"a}schel}, {et~al\mbox{.}}} \bibinfo{year}{2020}\natexlab{}.
\newblock \showarticletitle{Retrieval-augmented generation for knowledge-intensive nlp tasks}.
\newblock \bibinfo{journal}{\emph{Advances in Neural Information Processing Systems}}  \bibinfo{volume}{33} (\bibinfo{year}{2020}), \bibinfo{pages}{9459--9474}.
\newblock


\bibitem[\protect\citeauthoryear{Li, Li, Savarese, and Hoi}{Li et~al\mbox{.}}{2023b}]%
        {li2023blip}
\bibfield{author}{\bibinfo{person}{Junnan Li}, \bibinfo{person}{Dongxu Li}, \bibinfo{person}{Silvio Savarese}, {and} \bibinfo{person}{Steven Hoi}.} \bibinfo{year}{2023}\natexlab{b}.
\newblock \showarticletitle{Blip-2: Bootstrapping language-image pre-training with frozen image encoders and large language models}. In \bibinfo{booktitle}{\emph{International conference on machine learning}}. PMLR, \bibinfo{pages}{19730--19742}.
\newblock


\bibitem[\protect\citeauthoryear{Li, He, Wang, Li, Wang, Luo, Wang, Wang, and Qiao}{Li et~al\mbox{.}}{2023a}]%
        {li2023videochat}
\bibfield{author}{\bibinfo{person}{KunChang Li}, \bibinfo{person}{Yinan He}, \bibinfo{person}{Yi Wang}, \bibinfo{person}{Yizhuo Li}, \bibinfo{person}{Wenhai Wang}, \bibinfo{person}{Ping Luo}, \bibinfo{person}{Yali Wang}, \bibinfo{person}{Limin Wang}, {and} \bibinfo{person}{Yu Qiao}.} \bibinfo{year}{2023}\natexlab{a}.
\newblock \showarticletitle{Videochat: Chat-centric video understanding}.
\newblock \bibinfo{journal}{\emph{arXiv preprint arXiv:2305.06355}} (\bibinfo{year}{2023}).
\newblock


\bibitem[\protect\citeauthoryear{Li, Su, Zhang, Lin, and Li}{Li et~al\mbox{.}}{2015}]%
        {li2015trend}
\bibfield{author}{\bibinfo{person}{Li Li}, \bibinfo{person}{Xiaonan Su}, \bibinfo{person}{Yi Zhang}, \bibinfo{person}{Yuetong Lin}, {and} \bibinfo{person}{Zhiheng Li}.} \bibinfo{year}{2015}\natexlab{}.
\newblock \showarticletitle{Trend modeling for traffic time series analysis: An integrated study}.
\newblock \bibinfo{journal}{\emph{IEEE Transactions on Intelligent Transportation Systems}} \bibinfo{volume}{16}, \bibinfo{number}{6} (\bibinfo{year}{2015}), \bibinfo{pages}{3430--3439}.
\newblock


\bibitem[\protect\citeauthoryear{Li, Zhao, Li, Lu, Wang, Chang, Nie, Cao, Zhang, Sui, et~al\mbox{.}}{Li et~al\mbox{.}}{2022a}]%
        {li2022actionable}
\bibfield{author}{\bibinfo{person}{Zeyan Li}, \bibinfo{person}{Nengwen Zhao}, \bibinfo{person}{Mingjie Li}, \bibinfo{person}{Xianglin Lu}, \bibinfo{person}{Lixin Wang}, \bibinfo{person}{Dongdong Chang}, \bibinfo{person}{Xiaohui Nie}, \bibinfo{person}{Li Cao}, \bibinfo{person}{Wenchi Zhang}, \bibinfo{person}{Kaixin Sui}, {et~al\mbox{.}}} \bibinfo{year}{2022}\natexlab{a}.
\newblock \showarticletitle{Actionable and interpretable fault localization for recurring failures in online service systems}. In \bibinfo{booktitle}{\emph{Proceedings of the 30th ACM Joint European Software Engineering Conference and Symposium on the Foundations of Software Engineering}}. \bibinfo{pages}{996--1008}.
\newblock


\bibitem[\protect\citeauthoryear{Li, Zhao, Zhang, Sun, Chen, Wen, Ma, and Pei}{Li et~al\mbox{.}}{2022b}]%
        {li2022constructing}
\bibfield{author}{\bibinfo{person}{Zeyan Li}, \bibinfo{person}{Nengwen Zhao}, \bibinfo{person}{Shenglin Zhang}, \bibinfo{person}{Yongqian Sun}, \bibinfo{person}{Pengfei Chen}, \bibinfo{person}{Xidao Wen}, \bibinfo{person}{Minghua Ma}, {and} \bibinfo{person}{Dan Pei}.} \bibinfo{year}{2022}\natexlab{b}.
\newblock \showarticletitle{Constructing large-scale real-world benchmark datasets for aiops}.
\newblock \bibinfo{journal}{\emph{arXiv preprint arXiv:2208.03938}} (\bibinfo{year}{2022}).
\newblock


\bibitem[\protect\citeauthoryear{Lim and Zohren}{Lim and Zohren}{2021}]%
        {lim2021time}
\bibfield{author}{\bibinfo{person}{Bryan Lim} {and} \bibinfo{person}{Stefan Zohren}.} \bibinfo{year}{2021}\natexlab{}.
\newblock \showarticletitle{Time-series forecasting with deep learning: a survey}.
\newblock \bibinfo{journal}{\emph{Philosophical Transactions of the Royal Society A}} \bibinfo{volume}{379}, \bibinfo{number}{2194} (\bibinfo{year}{2021}), \bibinfo{pages}{20200209}.
\newblock


\bibitem[\protect\citeauthoryear{Liu, Li, Wu, and Lee}{Liu et~al\mbox{.}}{2024a}]%
        {liu2024visual}
\bibfield{author}{\bibinfo{person}{Haotian Liu}, \bibinfo{person}{Chunyuan Li}, \bibinfo{person}{Qingyang Wu}, {and} \bibinfo{person}{Yong~Jae Lee}.} \bibinfo{year}{2024}\natexlab{a}.
\newblock \showarticletitle{Visual instruction tuning}.
\newblock \bibinfo{journal}{\emph{Advances in neural information processing systems}}  \bibinfo{volume}{36} (\bibinfo{year}{2024}).
\newblock


\bibitem[\protect\citeauthoryear{Liu, Xu, Zhao, Kong, Kamarthi, Sasanur, Sharma, Cui, Wen, Zhang, et~al\mbox{.}}{Liu et~al\mbox{.}}{2024b}]%
        {liu2024time}
\bibfield{author}{\bibinfo{person}{Haoxin Liu}, \bibinfo{person}{Shangqing Xu}, \bibinfo{person}{Zhiyuan Zhao}, \bibinfo{person}{Lingkai Kong}, \bibinfo{person}{Harshavardhan Kamarthi}, \bibinfo{person}{Aditya~B Sasanur}, \bibinfo{person}{Megha Sharma}, \bibinfo{person}{Jiaming Cui}, \bibinfo{person}{Qingsong Wen}, \bibinfo{person}{Chao Zhang}, {et~al\mbox{.}}} \bibinfo{year}{2024}\natexlab{b}.
\newblock \showarticletitle{Time-MMD: A New Multi-Domain Multimodal Dataset for Time Series Analysis}.
\newblock \bibinfo{journal}{\emph{arXiv preprint arXiv:2406.08627}} (\bibinfo{year}{2024}).
\newblock


\bibitem[\protect\citeauthoryear{Luo, Cheng, Wang, Xu, Ni, Yu, Zhang, Liu, Chen, Chen, et~al\mbox{.}}{Luo et~al\mbox{.}}{2023}]%
        {luo2023time}
\bibfield{author}{\bibinfo{person}{Dongsheng Luo}, \bibinfo{person}{Wei Cheng}, \bibinfo{person}{Yingheng Wang}, \bibinfo{person}{Dongkuan Xu}, \bibinfo{person}{Jingchao Ni}, \bibinfo{person}{Wenchao Yu}, \bibinfo{person}{Xuchao Zhang}, \bibinfo{person}{Yanchi Liu}, \bibinfo{person}{Yuncong Chen}, \bibinfo{person}{Haifeng Chen}, {et~al\mbox{.}}} \bibinfo{year}{2023}\natexlab{}.
\newblock \showarticletitle{Time series contrastive learning with information-aware augmentations}. In \bibinfo{booktitle}{\emph{Proceedings of the AAAI Conference on Artificial Intelligence}}, Vol.~\bibinfo{volume}{37}. \bibinfo{pages}{4534--4542}.
\newblock


\bibitem[\protect\citeauthoryear{Luo, Zhang, Chen, Lin, Liu, Wu, Yang, Wang, Zeng, Gao, et~al\mbox{.}}{Luo et~al\mbox{.}}{2024}]%
        {luo2024mmevol}
\bibfield{author}{\bibinfo{person}{Run Luo}, \bibinfo{person}{Haonan Zhang}, \bibinfo{person}{Longze Chen}, \bibinfo{person}{Ting-En Lin}, \bibinfo{person}{Xiong Liu}, \bibinfo{person}{Yuchuan Wu}, \bibinfo{person}{Min Yang}, \bibinfo{person}{Minzheng Wang}, \bibinfo{person}{Pengpeng Zeng}, \bibinfo{person}{Lianli Gao}, {et~al\mbox{.}}} \bibinfo{year}{2024}\natexlab{}.
\newblock \showarticletitle{Mmevol: Empowering multimodal large language models with evol-instruct}.
\newblock \bibinfo{journal}{\emph{arXiv preprint arXiv:2409.05840}} (\bibinfo{year}{2024}).
\newblock


\bibitem[\protect\citeauthoryear{Maaz, Rasheed, Khan, and Khan}{Maaz et~al\mbox{.}}{2023}]%
        {maaz2023video}
\bibfield{author}{\bibinfo{person}{Muhammad Maaz}, \bibinfo{person}{Hanoona Rasheed}, \bibinfo{person}{Salman Khan}, {and} \bibinfo{person}{Fahad~Shahbaz Khan}.} \bibinfo{year}{2023}\natexlab{}.
\newblock \showarticletitle{Video-chatgpt: Towards detailed video understanding via large vision and language models}.
\newblock \bibinfo{journal}{\emph{arXiv preprint arXiv:2306.05424}} (\bibinfo{year}{2023}).
\newblock


\bibitem[\protect\citeauthoryear{Ma{\^\i}tre, Chemli, Chevalier, Dousset, Gitto, and Teste}{Ma{\^\i}tre et~al\mbox{.}}{2020}]%
        {maitre2020event}
\bibfield{author}{\bibinfo{person}{Elliot Ma{\^\i}tre}, \bibinfo{person}{Zakaria Chemli}, \bibinfo{person}{Max Chevalier}, \bibinfo{person}{Bernard Dousset}, \bibinfo{person}{Jean-Philippe Gitto}, {and} \bibinfo{person}{Olivier Teste}.} \bibinfo{year}{2020}\natexlab{}.
\newblock \showarticletitle{Event detection and time series alignment to improve stock market forecasting}. In \bibinfo{booktitle}{\emph{Joint conference of the information retrieval communities in europe (circle 2020)}}, Vol.~\bibinfo{volume}{2621}. CEUR-WS. org, \bibinfo{pages}{1--5}.
\newblock


\bibitem[\protect\citeauthoryear{Merrill, Tan, Gupta, Hartvigsen, and Althoff}{Merrill et~al\mbox{.}}{2024}]%
        {merrill2024language}
\bibfield{author}{\bibinfo{person}{Mike~A Merrill}, \bibinfo{person}{Mingtian Tan}, \bibinfo{person}{Vinayak Gupta}, \bibinfo{person}{Tom Hartvigsen}, {and} \bibinfo{person}{Tim Althoff}.} \bibinfo{year}{2024}\natexlab{}.
\newblock \showarticletitle{Language Models Still Struggle to Zero-shot Reason about Time Series}.
\newblock \bibinfo{journal}{\emph{arXiv preprint arXiv:2404.11757}} (\bibinfo{year}{2024}).
\newblock


\bibitem[\protect\citeauthoryear{Oh, Lee, Bae, Kwon, and Choi}{Oh et~al\mbox{.}}{2024}]%
        {oh2024ecg}
\bibfield{author}{\bibinfo{person}{Jungwoo Oh}, \bibinfo{person}{Gyubok Lee}, \bibinfo{person}{Seongsu Bae}, \bibinfo{person}{Joon-myoung Kwon}, {and} \bibinfo{person}{Edward Choi}.} \bibinfo{year}{2024}\natexlab{}.
\newblock \showarticletitle{Ecg-qa: A comprehensive question answering dataset combined with electrocardiogram}.
\newblock \bibinfo{journal}{\emph{Advances in Neural Information Processing Systems}}  \bibinfo{volume}{36} (\bibinfo{year}{2024}).
\newblock


\bibitem[\protect\citeauthoryear{Pan, Luo, Wang, Chen, Wang, and Wu}{Pan et~al\mbox{.}}{2024}]%
        {pan2024unifying}
\bibfield{author}{\bibinfo{person}{Shirui Pan}, \bibinfo{person}{Linhao Luo}, \bibinfo{person}{Yufei Wang}, \bibinfo{person}{Chen Chen}, \bibinfo{person}{Jiapu Wang}, {and} \bibinfo{person}{Xindong Wu}.} \bibinfo{year}{2024}\natexlab{}.
\newblock \showarticletitle{Unifying large language models and knowledge graphs: A roadmap}.
\newblock \bibinfo{journal}{\emph{IEEE Transactions on Knowledge and Data Engineering}} (\bibinfo{year}{2024}).
\newblock


\bibitem[\protect\citeauthoryear{Penfold and Zhang}{Penfold and Zhang}{2013}]%
        {penfold2013use}
\bibfield{author}{\bibinfo{person}{Robert~B Penfold} {and} \bibinfo{person}{Fang Zhang}.} \bibinfo{year}{2013}\natexlab{}.
\newblock \showarticletitle{Use of interrupted time series analysis in evaluating health care quality improvements}.
\newblock \bibinfo{journal}{\emph{Academic pediatrics}} \bibinfo{volume}{13}, \bibinfo{number}{6} (\bibinfo{year}{2013}), \bibinfo{pages}{S38--S44}.
\newblock


\bibitem[\protect\citeauthoryear{RB}{RB}{1990}]%
        {rb1990stl}
\bibfield{author}{\bibinfo{person}{CLEVELAND RB}.} \bibinfo{year}{1990}\natexlab{}.
\newblock \showarticletitle{STL: A seasonal-trend decomposition procedure based on loess}.
\newblock \bibinfo{journal}{\emph{J Off Stat}}  \bibinfo{volume}{6} (\bibinfo{year}{1990}), \bibinfo{pages}{3--73}.
\newblock


\bibitem[\protect\citeauthoryear{Rubenstein, Asawaroengchai, Nguyen, Bapna, Borsos, Quitry, Chen, Badawy, Han, Kharitonov, et~al\mbox{.}}{Rubenstein et~al\mbox{.}}{2023}]%
        {rubenstein2023audiopalm}
\bibfield{author}{\bibinfo{person}{Paul~K Rubenstein}, \bibinfo{person}{Chulayuth Asawaroengchai}, \bibinfo{person}{Duc~Dung Nguyen}, \bibinfo{person}{Ankur Bapna}, \bibinfo{person}{Zal{\'a}n Borsos}, \bibinfo{person}{F{\'e}lix de~Chaumont Quitry}, \bibinfo{person}{Peter Chen}, \bibinfo{person}{Dalia~El Badawy}, \bibinfo{person}{Wei Han}, \bibinfo{person}{Eugene Kharitonov}, {et~al\mbox{.}}} \bibinfo{year}{2023}\natexlab{}.
\newblock \showarticletitle{Audiopalm: A large language model that can speak and listen}.
\newblock \bibinfo{journal}{\emph{arXiv preprint arXiv:2306.12925}} (\bibinfo{year}{2023}).
\newblock


\bibitem[\protect\citeauthoryear{Savage}{Savage}{2023}]%
        {savage2023synthetic}
\bibfield{author}{\bibinfo{person}{Neil Savage}.} \bibinfo{year}{2023}\natexlab{}.
\newblock \showarticletitle{Synthetic data could be better than real data.}
\newblock \bibinfo{journal}{\emph{Nature}} (\bibinfo{year}{2023}).
\newblock


\bibitem[\protect\citeauthoryear{Sezer, Gudelek, and Ozbayoglu}{Sezer et~al\mbox{.}}{2020}]%
        {sezer2020financial}
\bibfield{author}{\bibinfo{person}{Omer~Berat Sezer}, \bibinfo{person}{Mehmet~Ugur Gudelek}, {and} \bibinfo{person}{Ahmet~Murat Ozbayoglu}.} \bibinfo{year}{2020}\natexlab{}.
\newblock \showarticletitle{Financial time series forecasting with deep learning: A systematic literature review: 2005--2019}.
\newblock \bibinfo{journal}{\emph{Applied soft computing}}  \bibinfo{volume}{90} (\bibinfo{year}{2020}), \bibinfo{pages}{106181}.
\newblock


\bibitem[\protect\citeauthoryear{Su, Jiang, Jin, Qiao, Xiao, Ma, Wei, Jing, Xu, and Lin}{Su et~al\mbox{.}}{2024}]%
        {su2024large}
\bibfield{author}{\bibinfo{person}{Jing Su}, \bibinfo{person}{Chufeng Jiang}, \bibinfo{person}{Xin Jin}, \bibinfo{person}{Yuxin Qiao}, \bibinfo{person}{Tingsong Xiao}, \bibinfo{person}{Hongda Ma}, \bibinfo{person}{Rong Wei}, \bibinfo{person}{Zhi Jing}, \bibinfo{person}{Jiajun Xu}, {and} \bibinfo{person}{Junhong Lin}.} \bibinfo{year}{2024}\natexlab{}.
\newblock \showarticletitle{Large language models for forecasting and anomaly detection: A systematic literature review}.
\newblock \bibinfo{journal}{\emph{arXiv preprint arXiv:2402.10350}} (\bibinfo{year}{2024}).
\newblock


\bibitem[\protect\citeauthoryear{Wang, Liu, Zhang, Zhong, Wang, Yin, Fan, Wu, and Wen}{Wang et~al\mbox{.}}{2024}]%
        {rcagent}
\bibfield{author}{\bibinfo{person}{Zefan Wang}, \bibinfo{person}{Zichuan Liu}, \bibinfo{person}{Yingying Zhang}, \bibinfo{person}{Aoxiao Zhong}, \bibinfo{person}{Jihong Wang}, \bibinfo{person}{Fengbin Yin}, \bibinfo{person}{Lunting Fan}, \bibinfo{person}{Lingfei Wu}, {and} \bibinfo{person}{Qingsong Wen}.} \bibinfo{year}{2024}\natexlab{}.
\newblock \showarticletitle{Rcagent: Cloud root cause analysis by autonomous agents with tool-augmented large language models}. In \bibinfo{booktitle}{\emph{Proceedings of the 33rd ACM International Conference on Information and Knowledge Management}}. \bibinfo{pages}{4966--4974}.
\newblock


\bibitem[\protect\citeauthoryear{Wolde-Rufael}{Wolde-Rufael}{2006}]%
        {wolde2006electricity}
\bibfield{author}{\bibinfo{person}{Yemane Wolde-Rufael}.} \bibinfo{year}{2006}\natexlab{}.
\newblock \showarticletitle{Electricity consumption and economic growth: a time series experience for 17 African countries}.
\newblock \bibinfo{journal}{\emph{Energy policy}} \bibinfo{volume}{34}, \bibinfo{number}{10} (\bibinfo{year}{2006}), \bibinfo{pages}{1106--1114}.
\newblock


\bibitem[\protect\citeauthoryear{Xing, Garcia, Cerutti, Kaplan, Preece, and Srivastava}{Xing et~al\mbox{.}}{2021}]%
        {xing2021deepsqa}
\bibfield{author}{\bibinfo{person}{Tianwei Xing}, \bibinfo{person}{Luis Garcia}, \bibinfo{person}{Federico Cerutti}, \bibinfo{person}{Lance Kaplan}, \bibinfo{person}{Alun Preece}, {and} \bibinfo{person}{Mani Srivastava}.} \bibinfo{year}{2021}\natexlab{}.
\newblock \showarticletitle{Deepsqa: Understanding sensor data via question answering}. In \bibinfo{booktitle}{\emph{Proceedings of the International Conference on Internet-of-Things Design and Implementation}}. \bibinfo{pages}{106--118}.
\newblock


\bibitem[\protect\citeauthoryear{Xu, Sun, Zheng, Geng, Zhao, Feng, Tao, and Jiang}{Xu et~al\mbox{.}}{2023}]%
        {xu2023wizardlm}
\bibfield{author}{\bibinfo{person}{Can Xu}, \bibinfo{person}{Qingfeng Sun}, \bibinfo{person}{Kai Zheng}, \bibinfo{person}{Xiubo Geng}, \bibinfo{person}{Pu Zhao}, \bibinfo{person}{Jiazhan Feng}, \bibinfo{person}{Chongyang Tao}, {and} \bibinfo{person}{Daxin Jiang}.} \bibinfo{year}{2023}\natexlab{}.
\newblock \showarticletitle{Wizardlm: Empowering large language models to follow complex instructions}.
\newblock \bibinfo{journal}{\emph{arXiv preprint arXiv:2304.12244}} (\bibinfo{year}{2023}).
\newblock


\bibitem[\protect\citeauthoryear{Yang, Yang, Hui, Zheng, Yu, Zhou, Li, Li, Liu, Huang, et~al\mbox{.}}{Yang et~al\mbox{.}}{2024}]%
        {yang2024qwen2}
\bibfield{author}{\bibinfo{person}{An Yang}, \bibinfo{person}{Baosong Yang}, \bibinfo{person}{Binyuan Hui}, \bibinfo{person}{Bo Zheng}, \bibinfo{person}{Bowen Yu}, \bibinfo{person}{Chang Zhou}, \bibinfo{person}{Chengpeng Li}, \bibinfo{person}{Chengyuan Li}, \bibinfo{person}{Dayiheng Liu}, \bibinfo{person}{Fei Huang}, {et~al\mbox{.}}} \bibinfo{year}{2024}\natexlab{}.
\newblock \showarticletitle{Qwen2 technical report}.
\newblock \bibinfo{journal}{\emph{arXiv preprint arXiv:2407.10671}} (\bibinfo{year}{2024}).
\newblock


\bibitem[\protect\citeauthoryear{Yao, Zhao, Yu, Du, Shafran, Narasimhan, and Cao}{Yao et~al\mbox{.}}{2022}]%
        {yao2022react}
\bibfield{author}{\bibinfo{person}{Shunyu Yao}, \bibinfo{person}{Jeffrey Zhao}, \bibinfo{person}{Dian Yu}, \bibinfo{person}{Nan Du}, \bibinfo{person}{Izhak Shafran}, \bibinfo{person}{Karthik Narasimhan}, {and} \bibinfo{person}{Yuan Cao}.} \bibinfo{year}{2022}\natexlab{}.
\newblock \showarticletitle{React: Synergizing reasoning and acting in language models}.
\newblock \bibinfo{journal}{\emph{arXiv preprint arXiv:2210.03629}} (\bibinfo{year}{2022}).
\newblock


\bibitem[\protect\citeauthoryear{Yin, Fu, Zhao, Li, Sun, Xu, and Chen}{Yin et~al\mbox{.}}{2024}]%
        {yin2024survey}
\bibfield{author}{\bibinfo{person}{Shukang Yin}, \bibinfo{person}{Chaoyou Fu}, \bibinfo{person}{Sirui Zhao}, \bibinfo{person}{Ke Li}, \bibinfo{person}{Xing Sun}, \bibinfo{person}{Tong Xu}, {and} \bibinfo{person}{Enhong Chen}.} \bibinfo{year}{2024}\natexlab{}.
\newblock \showarticletitle{A survey on multimodal large language models}.
\newblock \bibinfo{journal}{\emph{National Science Review}} (\bibinfo{year}{2024}), \bibinfo{pages}{nwae403}.
\newblock


\bibitem[\protect\citeauthoryear{Yoffe, Amayuelas, and Wang}{Yoffe et~al\mbox{.}}{2024}]%
        {yoffe2024debunc}
\bibfield{author}{\bibinfo{person}{Luke Yoffe}, \bibinfo{person}{Alfonso Amayuelas}, {and} \bibinfo{person}{William~Yang Wang}.} \bibinfo{year}{2024}\natexlab{}.
\newblock \showarticletitle{DebUnc: mitigating hallucinations in large language model agent communication with uncertainty estimations}.
\newblock \bibinfo{journal}{\emph{arXiv preprint arXiv:2407.06426}} (\bibinfo{year}{2024}).
\newblock


\bibitem[\protect\citeauthoryear{Yu, Guo, and Sano}{Yu et~al\mbox{.}}{2023}]%
        {yu2023zero}
\bibfield{author}{\bibinfo{person}{Han Yu}, \bibinfo{person}{Peikun Guo}, {and} \bibinfo{person}{Akane Sano}.} \bibinfo{year}{2023}\natexlab{}.
\newblock \showarticletitle{Zero-shot ECG diagnosis with large language models and retrieval-augmented generation}. In \bibinfo{booktitle}{\emph{Machine Learning for Health (ML4H)}}. PMLR, \bibinfo{pages}{650--663}.
\newblock


\bibitem[\protect\citeauthoryear{Zhang, Kuppannagari, Kannan, and Prasanna}{Zhang et~al\mbox{.}}{2018}]%
        {zhang2018generative}
\bibfield{author}{\bibinfo{person}{Chi Zhang}, \bibinfo{person}{Sanmukh~R Kuppannagari}, \bibinfo{person}{Rajgopal Kannan}, {and} \bibinfo{person}{Viktor~K Prasanna}.} \bibinfo{year}{2018}\natexlab{}.
\newblock \showarticletitle{Generative adversarial network for synthetic time series data generation in smart grids}. In \bibinfo{booktitle}{\emph{2018 IEEE international conference on communications, control, and computing technologies for smart grids (SmartGridComm)}}. IEEE, \bibinfo{pages}{1--6}.
\newblock


\bibitem[\protect\citeauthoryear{Zhang, Yu, Li, Dong, Su, Chu, and Yu}{Zhang et~al\mbox{.}}{2024b}]%
        {zhang2024mm}
\bibfield{author}{\bibinfo{person}{Duzhen Zhang}, \bibinfo{person}{Yahan Yu}, \bibinfo{person}{Chenxing Li}, \bibinfo{person}{Jiahua Dong}, \bibinfo{person}{Dan Su}, \bibinfo{person}{Chenhui Chu}, {and} \bibinfo{person}{Dong Yu}.} \bibinfo{year}{2024}\natexlab{b}.
\newblock \showarticletitle{Mm-llms: Recent advances in multimodal large language models}.
\newblock \bibinfo{journal}{\emph{arXiv preprint arXiv:2401.13601}} (\bibinfo{year}{2024}).
\newblock


\bibitem[\protect\citeauthoryear{Zhang, Li, and Bing}{Zhang et~al\mbox{.}}{2023}]%
        {zhang2023video}
\bibfield{author}{\bibinfo{person}{Hang Zhang}, \bibinfo{person}{Xin Li}, {and} \bibinfo{person}{Lidong Bing}.} \bibinfo{year}{2023}\natexlab{}.
\newblock \showarticletitle{Video-llama: An instruction-tuned audio-visual language model for video understanding}.
\newblock \bibinfo{journal}{\emph{arXiv preprint arXiv:2306.02858}} (\bibinfo{year}{2023}).
\newblock


\bibitem[\protect\citeauthoryear{Zhang, Sun, Wang, Fan, Mo, Xu, Liu, Yang, and Shi}{Zhang et~al\mbox{.}}{2024a}]%
        {zhang2024graphtranslator}
\bibfield{author}{\bibinfo{person}{Mengmei Zhang}, \bibinfo{person}{Mingwei Sun}, \bibinfo{person}{Peng Wang}, \bibinfo{person}{Shen Fan}, \bibinfo{person}{Yanhu Mo}, \bibinfo{person}{Xiaoxiao Xu}, \bibinfo{person}{Hong Liu}, \bibinfo{person}{Cheng Yang}, {and} \bibinfo{person}{Chuan Shi}.} \bibinfo{year}{2024}\natexlab{a}.
\newblock \showarticletitle{GraphTranslator: Aligning Graph Model to Large Language Model for Open-ended Tasks}. In \bibinfo{booktitle}{\emph{Proceedings of the ACM on Web Conference 2024}}. \bibinfo{pages}{1003--1014}.
\newblock


\bibitem[\protect\citeauthoryear{Zheng, Zhang, Zhang, Ye, Luo, Feng, and Ma}{Zheng et~al\mbox{.}}{2024}]%
        {zheng2024llamafactory}
\bibfield{author}{\bibinfo{person}{Yaowei Zheng}, \bibinfo{person}{Richong Zhang}, \bibinfo{person}{Junhao Zhang}, \bibinfo{person}{Yanhan Ye}, \bibinfo{person}{Zheyan Luo}, \bibinfo{person}{Zhangchi Feng}, {and} \bibinfo{person}{Yongqiang Ma}.} \bibinfo{year}{2024}\natexlab{}.
\newblock \showarticletitle{LlamaFactory: Unified Efficient Fine-Tuning of 100+ Language Models}. In \bibinfo{booktitle}{\emph{Proceedings of the 62nd Annual Meeting of the Association for Computational Linguistics (Volume 3: System Demonstrations)}}. \bibinfo{publisher}{Association for Computational Linguistics}, \bibinfo{address}{Bangkok, Thailand}.
\newblock
\urldef\tempurl%
\url{http://arxiv.org/abs/2403.13372}
\showURL{%
\tempurl}


\bibitem[\protect\citeauthoryear{Zhong, Fan, Zhang, Ma, Zhang, Sun, Lin, Zhang, and Pei}{Zhong et~al\mbox{.}}{2023}]%
        {zhong2023survey}
\bibfield{author}{\bibinfo{person}{Zhenyu Zhong}, \bibinfo{person}{Qiliang Fan}, \bibinfo{person}{Jiacheng Zhang}, \bibinfo{person}{Minghua Ma}, \bibinfo{person}{Shenglin Zhang}, \bibinfo{person}{Yongqian Sun}, \bibinfo{person}{Qingwei Lin}, \bibinfo{person}{Yuzhi Zhang}, {and} \bibinfo{person}{Dan Pei}.} \bibinfo{year}{2023}\natexlab{}.
\newblock \showarticletitle{A Survey of Time Series Anomaly Detection Methods in the AIOps Domain}.
\newblock \bibinfo{journal}{\emph{arXiv preprint arXiv:2308.00393}} (\bibinfo{year}{2023}).
\newblock


\bibitem[\protect\citeauthoryear{Zhou, Niu, Sun, Jin, et~al\mbox{.}}{Zhou et~al\mbox{.}}{2023b}]%
        {zhou2023one}
\bibfield{author}{\bibinfo{person}{Tian Zhou}, \bibinfo{person}{Peisong Niu}, \bibinfo{person}{Liang Sun}, \bibinfo{person}{Rong Jin}, {et~al\mbox{.}}} \bibinfo{year}{2023}\natexlab{b}.
\newblock \showarticletitle{One fits all: Power general time series analysis by pretrained lm}.
\newblock \bibinfo{journal}{\emph{Advances in neural information processing systems}}  \bibinfo{volume}{36} (\bibinfo{year}{2023}), \bibinfo{pages}{43322--43355}.
\newblock


\bibitem[\protect\citeauthoryear{Zhou, Li, Sun, Liu, Chen, Wu, Liu, Feng, and Zeng}{Zhou et~al\mbox{.}}{2023a}]%
        {dbot}
\bibfield{author}{\bibinfo{person}{Xuanhe Zhou}, \bibinfo{person}{Guoliang Li}, \bibinfo{person}{Zhaoyan Sun}, \bibinfo{person}{Zhiyuan Liu}, \bibinfo{person}{Weize Chen}, \bibinfo{person}{Jianming Wu}, \bibinfo{person}{Jiesi Liu}, \bibinfo{person}{Ruohang Feng}, {and} \bibinfo{person}{Guoyang Zeng}.} \bibinfo{year}{2023}\natexlab{a}.
\newblock \showarticletitle{D-bot: Database diagnosis system using large language models}.
\newblock \bibinfo{journal}{\emph{arXiv preprint arXiv:2312.01454}} (\bibinfo{year}{2023}).
\newblock


\bibitem[\protect\citeauthoryear{Zhuang, Yan, Zhang, Wang, Zhang, and Gu}{Zhuang et~al\mbox{.}}{2024}]%
        {zhuang2024see}
\bibfield{author}{\bibinfo{person}{Jiaxin Zhuang}, \bibinfo{person}{Leon Yan}, \bibinfo{person}{Zhenwei Zhang}, \bibinfo{person}{Ruiqi Wang}, \bibinfo{person}{Jiawei Zhang}, {and} \bibinfo{person}{Yuantao Gu}.} \bibinfo{year}{2024}\natexlab{}.
\newblock \showarticletitle{See it, Think it, Sorted: Large Multimodal Models are Few-shot Time Series Anomaly Analyzers}.
\newblock \bibinfo{journal}{\emph{arXiv preprint arXiv:2411.02465}} (\bibinfo{year}{2024}).
\newblock


\end{thebibliography}
\end{document}